\documentclass{article}
\usepackage[preprint]{neurips_2025}

\usepackage[utf8]{inputenc} 
\usepackage[T1]{fontenc}    
\usepackage{times,latexsym}
\usepackage{graphicx} \graphicspath{{figures/}}
\usepackage{amsmath,amssymb,mathabx,mathtools,amsthm,nicefrac}
\usepackage[linesnumbered,ruled,vlined]{algorithm2e}
\usepackage{acronym}
\usepackage{enumitem}
\usepackage[pagebackref,breaklinks,colorlinks]{hyperref}
\usepackage{balance}
\usepackage{xspace}
\usepackage{setspace}
\usepackage[skip=3pt,font=small]{subcaption}
\usepackage[skip=3pt,font=small]{caption}
\usepackage[dvipsnames,svgnames,x11names,table]{xcolor}
\usepackage[capitalise,noabbrev,nameinlink]{cleveref}
\usepackage{booktabs,tabularx,colortbl,multirow,multicol,array,makecell,tabularray}
\usepackage{overpic,wrapfig}
\usepackage[misc]{ifsym}
\usepackage{pifont}
\usepackage{tikz}

\usepackage[ruled,vlined,linesnumbered]{algorithm2e}
\SetKwComment{Comment}{/* }{ */}
\SetKwInOut{Input}{Input}
\SetKwInOut{Output}{Output}

\makeatletter
\DeclareRobustCommand\onedot{\futurelet\@let@token\@onedot}
\def\@onedot{\ifx\@let@token.\else.\null\fi\xspace}
\def\eg{\emph{e.g}\onedot}

\def\etal{\emph{et al}\onedot}
\makeatother

\makeatletter
\renewcommand{\paragraph}{%
  \@startsection{paragraph}{4}%
  {\z@}{0ex \@plus 0ex \@minus 0ex}{-1em}%
  {\hskip\parindent\normalfont\normalsize\bfseries}%
}
\makeatother

\crefname{algorithm}{Alg.}{Algs.}
\Crefname{algocf}{Algorithm}{Algorithms}
\crefname{section}{Sec.}{Secs.}
\Crefname{section}{Section}{Sections}
\crefname{table}{Tab.}{Tabs.}
\Crefname{table}{Table}{Tables}
\crefname{figure}{Fig.}{Figs.}
\Crefname{figure}{Figure}{Figures}
\crefname{equation}{Eq.}{Eqs.}
\Crefname{equation}{Equation}{Equations}
\crefname{appendix}{Appx.}{Appxs.}
\Crefname{appendix}{Appendix}{Appendices}

\definecolor{gblue}{HTML}{4285F4}
\definecolor{gred}{HTML}{DB4437}
\definecolor{ggreen}{HTML}{0F9D58}

\acrodef{fnf}[FNF]{Fourier Neural Filter}
\acrodef{dbd}[DBD]{Dual Branch Design}
\acrodef{fno}[FNO]{Fourier Neural Operator}
\acrodef{ffn}[FFN]{Feed-Forward Network}
\acrodef{mae}[MAE]{Mean Absolute Error}
\acrodef{mse}[MSE]{Mean Squared Error}

\title{Multivariate Long-term Time Series Forecasting\\with Fourier Neural Filter}

\author{%
    \normalfont\setlength{\tabcolsep}{3pt}%
    \hspace{-1em}\begin{tabular}{cccc}
        \textbf{Chenheng Xu}$^{\,1,2,3}$ & \textbf{Dan Wu}$^{\,1}$ & \textbf{Yixin Zhu}$^{\,2,3,4\,\textrm{\Letter}}$ & \textbf{Ying Nian Wu}$^{\,1,\,\textrm{\Letter}}$ \\
        \texttt{\footnotesize{}c.xu@ucla.edu} & \texttt{\footnotesize{}wudan11@ucla.edu} & \texttt{\footnotesize{}yixin.zhu@pku.edu.cn} & \texttt{\footnotesize{}ywu@stat.ucla.edu}
    \end{tabular}
    \vspace{6pt}\\
    $^1$ Department of Statistics \& Data Science, UCLA\\
    $^2$ School of Psychological and Cognitive Sciences, Peking University\\
    $^3$ Institute for Artificial Intelligence, Peking University\\
    $^4$ Beijing Key Laboratory of Behavior and Mental Health, Peking University
    \vspace{3pt}\\
    \url{https://chenheng-xu.github.io/fnf-time-series/}
    \vspace{-18pt}
}

\begin{document}
\maketitle

\begin{abstract}
Multivariate long-term time series forecasting has been suffering from the challenge of capturing both temporal dependencies within variables and spatial correlations across variables simultaneously~\cite{qiu2024tfb}.
Current approaches predominantly repurpose backbones from natural language processing or computer vision (\eg, Transformers), which fail to adequately address the unique properties of time series (\eg, periodicity)~\cite{wang2025timemixer++}.
The research community lacks a dedicated backbone with temporal-specific inductive biases, instead relying on domain-agnostic backbones supplemented with auxiliary techniques (\eg, signal decomposition).
We introduce \ac{fnf} as the backbone and \ac{dbd} as the architecture to provide excellent learning capabilities and optimal learning pathways for spatio-temporal modeling, respectively.
Our theoretical analysis proves that \ac{fnf} unifies local time-domain and global frequency-domain information processing within a single backbone that extends naturally to spatial modeling, while information bottleneck theory demonstrates that \ac{dbd} provides superior gradient flow and representation capacity compared to existing unified or sequential architectures.
Our empirical evaluation across 11 public benchmark datasets spanning five domains (energy, meteorology, transportation, environment, and nature) confirms state-of-the-art performance with consistent hyperparameter settings.
Notably, our approach achieves these results without any auxiliary techniques, suggesting that properly designed neural architectures can capture the inherent properties of time series, potentially transforming time series modeling in scientific and industrial applications.
\end{abstract}

\section{Introduction}

\begin{wrapfigure}{r}{0.5\linewidth}
    \centering
    \vspace{-48pt}
    \includegraphics[width=\linewidth]{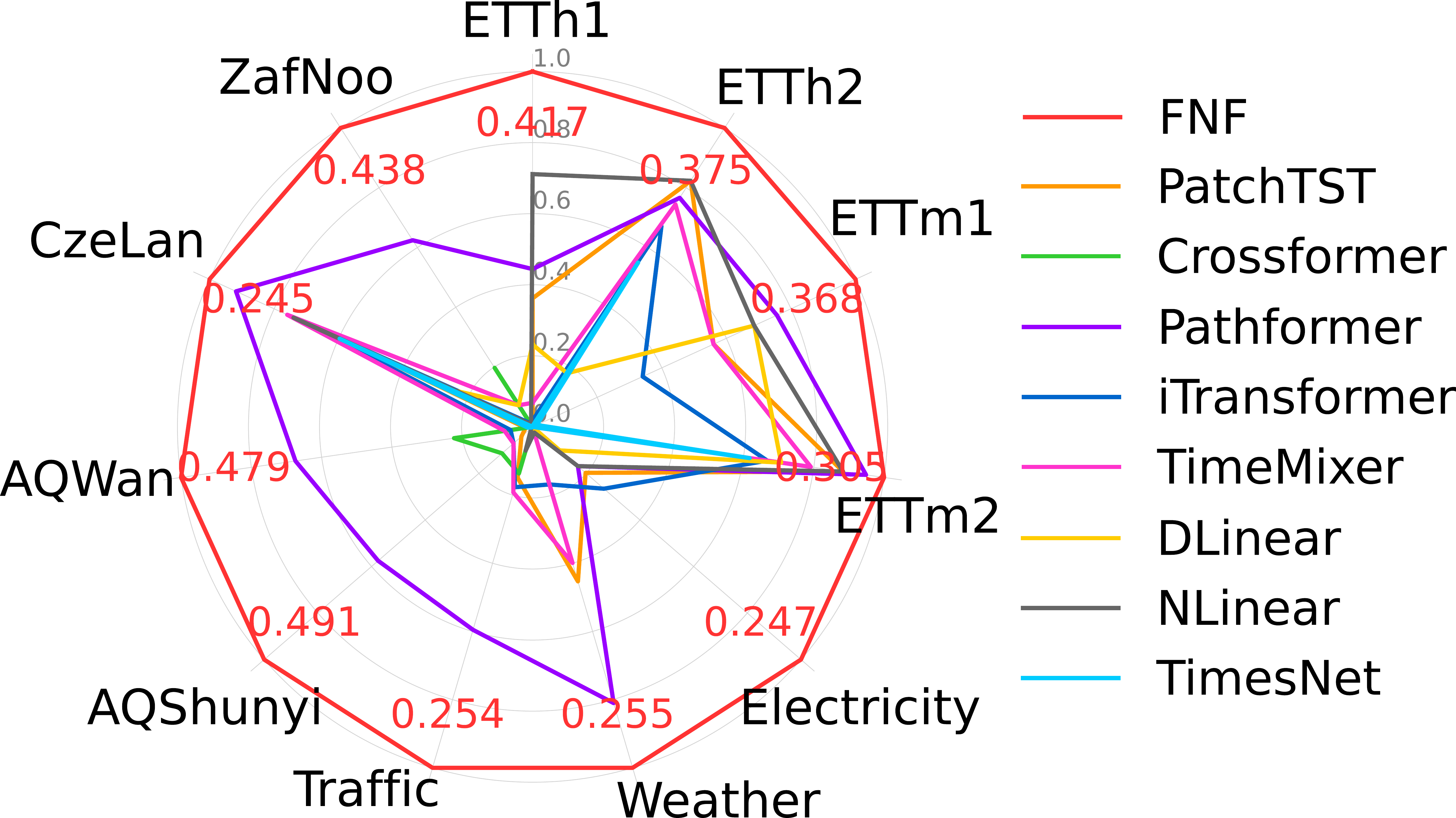}
    \caption{\textbf{Radar chart of forecasting performance across 11 benchmark datasets spanning five domains.} The chart displays average \acs{mae} across different forecast horizons of 96, 192, 336, 720. Our proposed \acs{fnf} (highlighted) consistently outperforms eight strong baseline models on diverse domains of energy, weather, transportation, environment, and nature with \textbf{consistent hyperparameter settings}.}
    \label{fig:radar_chart}
    \vspace{-18pt}
\end{wrapfigure}

Time series forecasting, which predicts future values based on historical observations, has attracted substantial academic attention and found widespread application across diverse domains, including energy~\cite{zhang2025machine} and meteorology~\cite{pathak2022fourcastnet,bi2023accurate}. Recent research has increasingly focused on multivariate long-term forecasting~\cite{zhou2021informer}, which presents several significant challenges. Extended forecast horizons inevitably increase uncertainty and degrade prediction accuracy, while complex temporal dependencies within variables and spatial correlations across variables further complicate modeling, particularly with high-dimensional variables. Consequently, developing neural architectures capable of simultaneously capturing heterogeneous temporal and spatial patterns has become critical for advancing time series forecasting.

For \textbf{within-variable} temporal modeling, the research community has developed several elaborate approaches. These include: (i) decomposing time series into trend, seasonal, and residual terms ~\cite{taylor2018forecasting,oreshkin2019n}; (ii) deconstructing time series into frequency components for \textit{multi-resolution modeling}~\cite{dai2024periodicity}; and (iii) segmenting sequences into patches of uniform~\cite{nie2022time,lee2023learning} or varying sizes~\cite{zhang2024multi,yu2024prformer} to capture both short-term fluctuations and long-term dependencies through \textit{multi-scale modeling}. More sophisticated techniques integrate multi-resolution and multi-scale modeling simultaneously~\cite{wang2025timemixer++}, establishing comprehensive frameworks for time series representation learning.

Despite these advances, existing approaches primarily rely on architectures borrowed from natural language processing or computer vision, such as Transformers~\cite{vaswani2017attention}, CNNs~\cite{luo2024moderntcn,wu2022timesnet}, and MLPs~\cite{zeng2023transformers}. These domain-agnostic architectures cannot fully address the inherent properties of time series without auxiliary techniques (\eg, signal decomposition). To address this fundamental limitation, we propose the \textbf{\acf{fnf}}, a novel nonlinear integral kernel operator that integrates temporal-specific inductive biases directly into the backbone design. Mathematically, \ac{fnf} extends the standard \ac{fno}~\cite{li2020fourier} by introducing an input-dependent kernel function that adaptively modulate information flow based on input properties. This design enables selective activation of local time-domain information and global frequency-domain information through Hadamard product operations, making it particularly effective for capturing the unique properties of time series. \ac{fnf} offers several key advantages: (i) it naturally extends to spatial modeling~\cite{guibas2021adaptive}; (ii) it achieves $O(N \log N)$ computational complexity compared to the $O(N^2)$ complexity of Transformers; and (iii) it internally incorporates linear expansion operations, eliminating the need for dedicated feed-forward networks and reducing parameter count while maintaining performance.

For \textbf{cross-variable} spatial modeling, the researchers have developed diverse techniques: (i) \textit{independent variable modeling}~\cite{nie2022time,dai2024periodicity}, which maintains stability but ignores inter-variable interactions; (ii) \textit{unified variable modeling}~\cite{zhang2023crossformer}, which comprehensively captures relationships but exhibits sensitivity to irrelevant variable disturbances; and (iii) \textit{hierarchical variable modeling}~\cite{chen2024similarity}, which provides a compromise approach but constrains flexibility by confining relationship patterns within predetermined cluster boundaries. These techniques highlight the fundamental trade-offs in spatial modeling and emphasize the need for adaptive systems that optimize across these competing priorities.

To address above challenges, we propose the \textbf{\acf{dbd}}. From an information bottleneck perspective~\cite{tishby2015deep}, this parallel dual-branch architecture optimizes information extraction and compression in multivariate time series by maintaining separate processing pathways for temporal and spatial patterns. Unlike unified techniques that suffer from the curse of dimensionality or sequential techniques~\cite{zhang2023crossformer,chen2024pathformer} that experience cascading information loss due to unequal information processing, \ac{dbd} ensures that each branch independently achieves optimal trade-offs between information extraction and compression while providing shorter and more direct gradient flow.

To validate our approach, we conducted comprehensive experiments across 11 datasets spanning five domains (energy, meteorology, transportation, environment, and nature). Our extensive evaluation demonstrates that our approach achieves state-of-the-art results, as shown in \cref{fig:radar_chart}.

\section{Related Work}

\paragraph{Distribution Shift}

The statistical properties of time series, such as mean and variance, tend to change over time, creating challenges for forecasting models~\cite{passalis2019deep,dai2024ddn,liu2025timebridge}. Researchers have developed various solutions to address this issue. RevIN~\cite{kim2021reversible} applies instance normalization on input sequences and performs de-normalization on output sequences. Dish-TS~\cite{fan2023dish} extracts distribution coefficients for both intra- and inter-variable dimensions to mitigate distribution shift. SAN~\cite{liu2023adaptive} addresses non-stationarity at the local temporal slice level using a lightweight network to model evolving statistical properties. FAN~\cite{ye2024frequency} handles dynamic trends and seasonal patterns by employing Fourier transform to identify key frequency components across instances. Notably, Li \etal~\cite{li2023revisiting} demonstrated that after using instance normalization, excellent results can be easily obtained with just a simple linear layer. Our approach employs basic instance normalization while introducing a backbone that inherently processes time-domain and frequency-domain information simultaneously.

\paragraph{Patch Embedding}

Time series patching has evolved from simple segmentation to sophisticated strategies that balance local and global information extraction~\cite{nie2022time,lee2023learning}. These techniques include overlapping~\cite{nie2022time,luo2024moderntcn} or non-overlapping~\cite{zhang2023crossformer} patching, variable length patching~\cite{yu2024prformer,zhang2024multi}, and hierarchical patching~\cite{huang2024hdmixer}. Recent research has explored adaptive patching strategies based on input properties of time series~\cite{chen2024pathformer}. While these approaches enhance representation capabilities, they primarily function as \textit{preprocessing} steps for conventional architectures. Our work adopts basic overlapping patching for fair comparison while introducing the \ac{fnf} backbone specifically designed to unify local time-domain and global frequency-domain processing for time series.

\paragraph{Non-Autoregressive}

To alleviate the error accumulation problem in autoregressive decoding, non-autoregressive approaches~\cite{wu2021autoformer,zhou2021informer,wu2022timesnet,liu2023itransformer} have become the standard paradigm for long-term time series forecasting. This technique simultaneously generates all future values through a linear layer, rather than recursively using previous predictions as inputs. While patch-based autoregressive models excel in large-scale time series foundation models~\cite{das2024decoder,shi2024time}, non-autoregressive approaches perform better in typical forecasting scenarios. Recent research has identified that non-autoregressive approaches implicitly assume conditional independence between future values, ignoring the temporal autocorrelation inherent in time series~\cite{wang2024fredf}. Our proposed \ac{fnf} and \ac{dbd} enhance representation learning quality within this established architecture, improving forecasting accuracy despite the theoretical limitations of the non-autoregressive paradigm.

\begin{wrapfigure}{r}{0.4\linewidth}
    \vspace{-54pt}
    \centering
    \includegraphics[width=\linewidth]{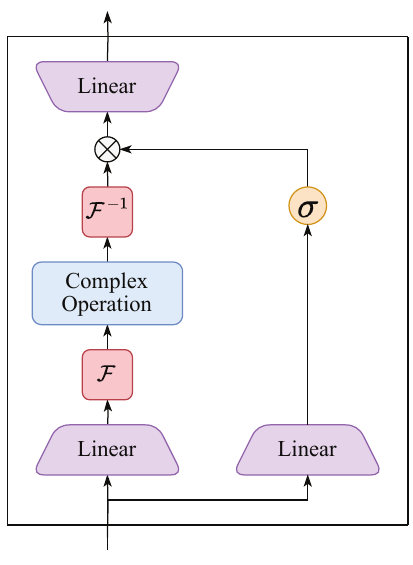}
    \caption{\textbf{The \acs{fnf} backbone.} Our dual-branch design processes input embeddings through parallel expanded linear layers~\cite{gu2024mamba}. The right branch captures time-domain patterns via GELU activation~\cite{hendrycks2016gaussian}, while the left branch extracts frequency-domain features through Fourier transform, complex operations~\cite{guibas2021adaptive} (two complex linear layers with Softshrink function~\cite{donoho2002noising}), and inverse Fourier transform. The branches are combined via Hadamard product ($\odot$), enabling simultaneous modeling of local temporal and global spectral information.}
    \label{fig:fnf_backbone}
    \vspace{-30pt}
\end{wrapfigure}

\section{Methodology}
\label{sec:methodology}

This section establishes the theoretical foundations of our proposed \acf{fnf} backbone and the \acf{dbd} architecture.

\subsection{The \acf{fnf} Backbone}

While \ac{fno}~\cite{li2020fourier} has demonstrated remarkable effectiveness in modeling complex dynamic systems and solving partial differential equations through fixed integration kernels, our proposed \ac{fnf} (\cref{fig:fnf_backbone}) introduces a critical advancement---an input-dependent integral kernel that significantly enhances flexibility and generalization capacity. We analyze the theoretical underpinnings of \ac{fnf} by examining integral kernel functions, global convolution operations, selective activation mechanisms, and their connections to Transformer architectures.

\subsubsection{Integral Kernel}

\paragraph{Definition 1} \ac{fno}~\cite{li2020fourier} is defined through an integral kernel operator:
\begin{equation}
    (K v)(x) = \int_D \kappa(x, y) v(y) \, dy,
\end{equation}
where $\kappa: D \times D \rightarrow \mathbb{R}$ is the kernel function and $v: D \rightarrow \mathbb{R}$ is the input function. Through the Fourier transform, \ac{fno} can be formulated in the frequency domain as:
\begin{equation}
    (K v)(x) = \mathcal{F}^{-1}(R_\phi \cdot \mathcal{F}(v))(x),
\end{equation}
where $R_\phi = \mathcal{F}(\kappa)$ denotes the parameterized frequency-domain kernel.

\paragraph{Definition 2} \ac{fnf} extends this concept through a generalized integral kernel operator:
\begin{equation}
    (K v)(x) = \int_D \kappa(x, y; v) v(y) \, dy,
\end{equation}
where the kernel function $\kappa(x, y; v)$ is input-dependent. In implementation, \ac{fnf} takes the form:
\begin{equation}
    (K v)(x) = T(G(v)(x) \odot \mathcal{F}^{-1}(R_\phi \cdot \mathcal{F}(H(v)))(x)),
\end{equation}
where $T(v)$, $G(v)$, and $H(v)$ represent linear transformations, with $G$ activated by GELU~\cite{hendrycks2016gaussian}, and $\odot$ denoting the Hadamard product operation.

\paragraph{Remark 1} The fundamental distinction between \ac{fno} and \ac{fnf} lies in their kernel functions: \ac{fno} employs a fixed kernel $\kappa(x, y)$, whereas \ac{fnf} utilizes an input-dependent kernel $\kappa(x,y;v)$, enabling adaptive information flow modulation based on input properties.

\subsubsection{Global Convolution}

\paragraph{Definition 3} When the kernel function $\kappa(x, y) = \kappa(x-y)$ exhibits translation invariance, the integral kernel operator in \ac{fno} reduces to a global convolution~\cite{li2020fourier}:
\begin{equation}
    (K v)(x) = \int_D \kappa(x-y) v(y) \, dy = (\kappa * v)(x).
\end{equation}

\paragraph{Definition 4} Similarly, when the kernel function $\kappa(x, y; v) = \kappa(x-y; v)$ maintains translation invariance, the integral kernel operator in \ac{fnf} becomes a gated global convolution:
\begin{equation}
    (K v)(x) = \int_D \tilde{\kappa}(x-y; v) v(y) \, dy = (\tilde{\kappa}(\cdot; v) * v)(x).
\end{equation}

\paragraph{Remark 2} Translation invariance provides a computational advantage for both \ac{fno} and \ac{fnf}, enabling efficient implementation through Fourier domain operations. However, the gated global convolution of \ac{fnf} introduces enhanced expressivity through its input-dependent kernel $\tilde{\kappa}(\cdot; v)$, which dynamically adapts its filtering behavior while preserving the computational efficiency of convolution operations.

\subsubsection{Selective Activation}

\paragraph{Definition 5} The selective activation mechanism in \ac{fnf} operates through Hadamard product between local information $G(v)(x)$ and global information $P(v)(x)$:
\begin{equation}
    (G(v) \odot P(v))_i = |G(v)_i| \cdot |P(v)_i| \cdot e^{i(\theta_{G(v)_i} + \theta_{P(v)_i})},
\end{equation}
where $|G(v)_i|$ and $|P(v)i|$ represent magnitudes and $\theta{G(v)i}$ and $\theta{P(v)_i}$ represent phases.

\paragraph{Remark 3} This formulation reveals how selective activation simultaneously modulates both amplitude (through multiplication $|G(v)_i| \cdot |P(v)_i|$) and phase (through addition $\theta_{G(v)_i} + \theta_{P(v)_i}$). This joint modulation in complex space enables precise control over both intensity and orientation of features, providing substantially enhanced representational capacity compared to conventional approaches.

\paragraph{Remark 4} In \ac{fnf}, selective activation facilitates adaptive fusion of local and global information. The local information $G(v)$ functions as a gating mechanism that regulates global information flow from $P(v)$ at each position. This creates three distinct operational modes: suppression when $|G(v)_i| \approx 0$, preservation when $|G(v)_i| \approx 1$, and amplification when $|G(v)_i| > 1$, allowing \ac{fnf} to dynamically balance local details with global context by focusing computational resources on the most informative regions.

\subsubsection{Complex Operation}

\paragraph{Definition 7} Complex linear transformation~\cite{guibas2021adaptive} operates on complex-valued input $z = z_r + iz_i$ with complex weights $W = W_r + iW_i$ and biases $b = b_r + ib_i$:
\begin{equation}
   L(z) = (W_r z_r - W_i z_i + b_r) + i(W_r z_i + W_i z_r + b_i).
\end{equation}

\paragraph{Remark 5} Complex activation functions typically apply standard nonlinear activations like ReLU~\cite{nair2010rectified} separately to real and imaginary components: $\sigma_\mathbb{C}(z) = \sigma(z_r) + i\sigma(z_i)$. In our implementation, we employ GELU~\cite{hendrycks2016gaussian} as the complex activation function.

\paragraph{Definition 8} The Softshrink operation~\cite{donoho2002noising} for frequency domain denoising is defined as:
\begin{equation}
   S_\lambda(z) = \begin{cases} 
   (|z| - \lambda)\frac{z}{|z|}, & \text{if } |z| > \lambda \\
   0, & \text{if } |z| \leq \lambda, 
   \end{cases}
\end{equation}
where $\lambda$ is the threshold parameter.

\paragraph{Remark 6} The Softshrink operation preserves phase while adaptively sparsifying the frequency spectrum by eliminating components below threshold $\lambda$, effectively filtering noise while maintaining significant frequency information.

\subsubsection{Connection to Transformers}

\paragraph{Functionality} \ac{fnf} represents a unified backbone that implements core Transformer functions through alternative computational mechanisms. The gated global convolution in \ac{fnf} performs comprehensive information exchange across all positions, analogous to token mixing in self-attention. Furthermore, the linear transformations ($T(v)$, $G(v)$, and $H(v)$) in \ac{fnf} can be expanded to replicate the functionality of feed-forward networks, establishing functional equivalence between architectures while maintaining distinct computational pathways.

\paragraph{Complexity} \ac{fnf} achieves token mixing with $O(N \log N)$ computational complexity through Fourier transform operations, compared to the $O(N^2)$ complexity of standard Transformers for sequence length $N$. Moreover, \ac{fnf} typically requires fewer parameters while maintaining comparable expressivity, making it particularly efficient for modeling long-range dependencies in high-dimensional temporal sequences.

\subsection{The \acf{dbd} Architecture}

\begin{wrapfigure}[34]{r}{0.38\linewidth}
    \vspace{-45pt}
    \centering
    \begin{subfigure}[t]{0.5\linewidth}
        \includegraphics[width=\linewidth]{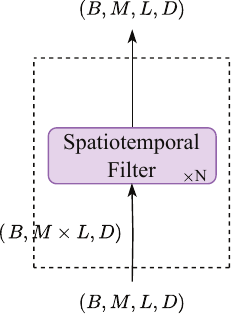}
        \caption{Unified}
    \end{subfigure}%
    \begin{subfigure}[t]{0.5\linewidth}
        \includegraphics[width=\linewidth]{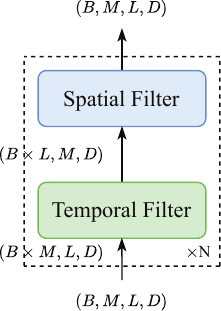}
        \caption{Sequential}
    \end{subfigure}%
    \\%
    \begin{subfigure}[t]{\linewidth}
        \includegraphics[width=\linewidth]{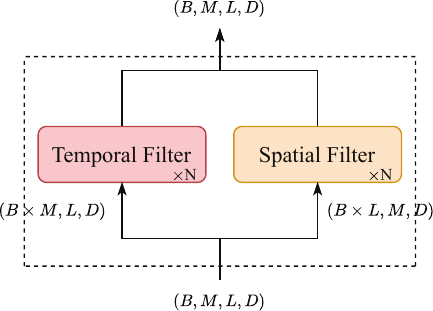}
        \caption{Parallel}
    \end{subfigure}%
    \caption{\textbf{The \acf{dbd} architecture.} Each approach processes multivariate time series in shape $(B, M, L, D)$ differently: (a) \textbf{Unified}: single backbone simultaneously models temporal and spatial patterns; (b) \textbf{Sequential}: cascades temporal filter followed by spatial filter; (c) \textbf{Parallel}: applies independent temporal and spatial filters in separate branches. Our \acs{dbd} implements the parallel design to effectively perform temporal and spatial modeling.}
    \label{fig:dbd_architecture}
\end{wrapfigure}

Through the lens of information bottleneck theory~\cite{tishby2015deep}, spatiotemporal modeling architectures can be categorized into three paradigms (\cref{fig:dbd_architecture}): unified (which suffers from the curse of dimensionality)~\cite{alemi2016deep}, sequential (which creates information bottlenecks)~\cite{zhang2023crossformer, chen2024pathformer}, and parallel (which preserves information through independent branches). Our \ac{dbd} architecture adopts the parallel approach to maximize gradient flow and representation capacity.

\paragraph{Unified} The unified paradigm captures temporal and spatial, and spatiotemporal patterns through a single operation, formulated as:
\begin{equation}
    X \rightarrow T \rightarrow Y,
\end{equation}
where $T$ represents a high-dimensional intermediate representation. As $T$ must simultaneously encode both temporal and spatial information, finding the optimal balance between compression and informativeness becomes computationally intractable. This approach requires substantially more parameters to achieve comparable expressiveness, increasing overfitting risk and reducing generalization performance.

\paragraph{Sequential} The sequential paradigm implements a cascaded architecture where temporal processing precedes spatial processing:
\begin{equation}
    X \rightarrow T_1 \rightarrow T_2 \rightarrow Y,
\end{equation}
where $T_1$ and $T_2$ denote temporal and spatial representations, respectively. According to the information processing inequality, this architecture inherently suffers from information loss:
\begin{equation}
    I(X ; Y) \geq I\left(T_1 ; Y\right) \geq I\left(T_2 ; Y\right).
\end{equation}
Each processing stage acts as an information bottleneck, potentially discarding critical temporal patterns before they reach the spatial processing stage. Additionally, gradients must propagate through multiple transformation layers, often leading to vanishing gradient problems.

\paragraph{Parallel} The parallel paradigm maintains independent information processing branches with direct access to the original input~\cite{wang2018non}:
\begin{equation}
    X \rightarrow\left\{
    \begin{array}{l}
        T_1 \\
        T_2
    \end{array} \rightarrow T \rightarrow Y\right..
\end{equation}
This approach ensures each branch independently achieves optimal information compression, with each optimizing its own objective function:
\begin{equation}
    \begin{aligned}
        & \min_{T_1}\left[I\left(T_1 ; X\right)-\beta_1 I\left(T_1 ; Y\right)\right] \\
        & \min_{T_2}\left[I\left(T_2 ; X\right)-\beta_2 I\left(T_2 ; Y\right)\right],
    \end{aligned}
\end{equation}
where $\beta_1$ and $\beta_2$ control the trade-off between compression and performance for each branch. This allows temporal and spatial processors to specialize without constraints imposed by the other branch.

\paragraph{Gradient Flow} The parallel architecture offers significant advantages in terms of gradient flow. With independent branches, the gradient paths remain short and direct:
\begin{equation}
    \begin{aligned}
        \frac{\partial L}{\partial \theta_1} & =\frac{\partial L}{\partial g} \cdot \frac{\partial g}{\partial f_1} \cdot \frac{\partial f_1}{\partial \theta_1} \\
        \frac{\partial L}{\partial \theta_2} & =\frac{\partial L}{\partial g} \cdot \frac{\partial g}{\partial f_2} \cdot \frac{\partial f_2}{\partial \theta_2}.
    \end{aligned}
\end{equation}
These shorter gradient paths significantly reduce the risk of vanishing or exploding gradients that typically plague deeper sequential architectures. Additionally, the independent branches enable more efficient parallelization during both forward and backward passes, accelerating convergence without sacrificing model expressiveness.

\paragraph{Representation Capacity} The parallel architecture demonstrates superior representation capacity through its ability to retain and complement information. The theoretical upper bound on mutual information is given by:
\begin{equation}
    I\left(T ; Y\right) \leq I\left(T_1, T_2 ; Y\right).
\end{equation}
Due to the complementary nature of temporal and spatial information, the joint representation typically satisfies:
\begin{equation}
    I\left(T_1, T_2 ; Y\right)>\max \left\{I\left(T_1 ; Y\right), I\left(T_2 ; Y\right)\right\}.
\end{equation}
This inequality holds because temporal and spatial features capture fundamentally orthogonal aspects of the input---temporal features extract dynamic patterns across time, while spatial features encode structural relationships within each time point. Their combination provides a more comprehensive view than either branch could achieve alone. Furthermore, the parallel approach enables adaptive information fusion through a learnable, input-dependent weights:
\begin{equation}
    T=\alpha(X) \cdot T_1+ (1-\alpha(X)) \cdot T_2.
\end{equation}
This dynamic weighting mechanism contextually assesses the relative importance of each information flow, effectively addressing varying requirements across different scenarios and input characteristics.

\section{Model Implementation}
\label{sec:model}

\paragraph{Overview} We denote the lookback window as $X=\left(x_1, \ldots, x_T\right) \in \mathbb{R}^{L \times M}$ with $L$ timesteps and $M$ variables, and the forecast horizon as $Y=\left(y_1, \ldots, y_L\right) \in \mathbb{R}^{H \times M}$ with $H$ timesteps. 

\begin{figure}[t!]
    \centering
    \includegraphics[width=\linewidth]{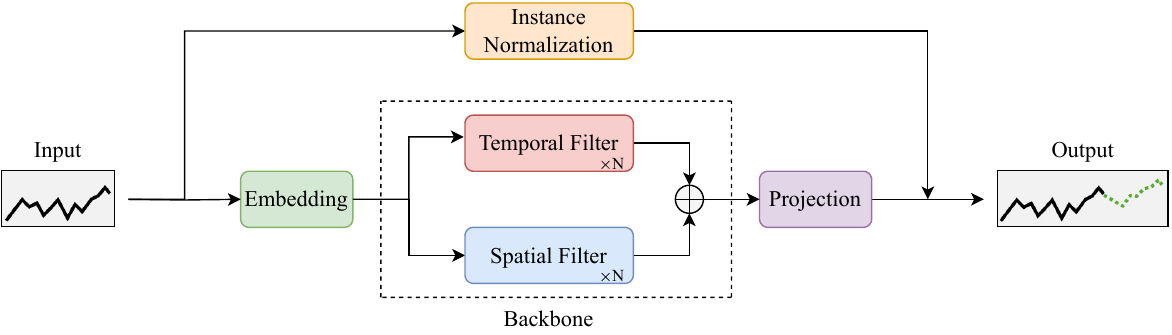}
    \caption{\textbf{Schematic diagram of our proposed neural architecture.} Where, normalization for mitigating distribution shift, embedding for obtaining informative input embeddings, backbone for addressing temporal dependencies and spatial correlations, and projection for generating predicted values.}
    \label{fig:model_schematic}
\end{figure}

As illustrated in \cref{fig:model_schematic}, our pipeline integrates specialized components---normalization, embedding, dual-branch backbone, projection, and de-normalization---each addressing specific aspects of the forecasting challenge.

\paragraph{Normalization} To address distribution shift inherent in the time series, we implement instance normalization~\cite{kim2021reversible}:
\begin{eqnarray}
    X = \frac{X - \mu}{\sigma + \epsilon},
\end{eqnarray}
where $\mu$ and $\sigma$ represent the mean and standard deviation vectors of the input sequence, with $\epsilon$ added for numerical stability. 

\paragraph{Embedding} We transform the normalized sequences using patch embedding~\cite{nie2022time} to enhance feature representation:
\begin{equation}
    X = \text{Patching}(X) + \text{PE},
\end{equation}
where $\text{Patching}$ segments time series into overlapping patches, and $\text{PE}$ incorporates sine-cosine positional encoding to preserve temporal ordering.

\paragraph{Backbone} The core design leverages a parallel dual-branch architecture with adaptive gated mechanism:
\begin{equation}
    X = \alpha \cdot \text{\ac{fnf}}_1(X) + (1-\alpha) \cdot \text{\ac{fnf}}_2(X),
\end{equation}
where $\text{\ac{fnf}}_1$ extracts temporal dependencies while $\text{\ac{fnf}}_2$ captures spatial correlations across variables. The dynamic gating coefficient $\alpha = \sigma(W \cdot \text{\ac{fnf}}_1(X) + b)$ adaptively balances these complementary information streams, with $\sigma$ representing the sigmoid activation~\cite{rumelhart1986learning}, and $W$ and $b$ as learnable parameters.

\paragraph{Projection} The processed embeddings are transformed into predictions through a straightforward linear projection:
\begin{eqnarray}
    Y = \text{Linear}(\text{Flatten}(X)),
\end{eqnarray}
where flattening and linear transformation efficiently map the embedded features to the normalized output space.

\paragraph{De-normalization} Finally, we restore the predictions to their original scale:
\begin{eqnarray}
    Y = Y \cdot (\sigma + \epsilon) + \mu,
\end{eqnarray}
applying the inverse of the normalization process to obtain future values.

\section{Experiments}
\label{sec:experiments}

\paragraph{Datasets} To thoroughly evaluate our model performance across diverse domains, we conduct experiments on 11 real-world datasets spanning 5 distinct domains, as detailed in \cref{tab:dataset_info}. These include energy (ETTh1, ETTh2, ETTm1, ETTm2, and Electricity)~\cite{zhou2021informer}, meteorology (Weather)~\cite{wu2021autoformer}, transportation (Traffic)~\cite{wu2021autoformer}, environment (AQShunyi and AQWan)~\cite{zhang2017cautionary}, and nature (ZafNoo and CzeLan)~\cite{poyatos2020global}.

\begin{wraptable}{r}{0.4\linewidth}
    \centering
    \vspace{-9pt}
    \caption{\textbf{Dataset overview.}}
    \label{tab:dataset_info}
    \resizebox{\linewidth}{!}{%
        \begin{tabular}{@{}ccccc@{}}
            \toprule
            Domain & Dataset & Frequency & Length & Variable \\
            \midrule
            \multirow{5}{*}{Energy} 
                & ETTh1 & 1 hour & 14,400 & 7 \\
                & ETTh2 & 1 hour & 14,400 & 7 \\
                & ETTm1 & 15 min & 57,600 & 7 \\
                & ETTm2 & 15 min & 57,600 & 7 \\
                & Electricity & 1 hour & 26,304 & 321 \\
            \midrule
            \multirow{1}{*}{Meteorology} 
                & Weather & 10 min & 52,696 & 21 \\
            \midrule
            \multirow{1}{*}{Transportation} 
                & Traffic & 1 hour & 17,544 & 862 \\
            \midrule
            \multirow{2}{*}{Environment} 
                & AQShunyi & 1 hour & 35,064 & 11 \\
                & AQWan & 1 hour & 35,064 & 11 \\
            \midrule
            \multirow{2}{*}{Nature} 
                & ZafNoo & 2 min & 19,225 & 11 \\
                & CzeLan & 30 min & 19,954 & 11 \\
            \bottomrule
        \end{tabular}%
    }%
    \vspace{-9pt}
\end{wraptable}

\paragraph{Baselines} We benchmark our approach against recent state-of-the-art models across three architectural paradigms: (i) Transformer-based models: Pathformer~\cite{chen2024pathformer}, PatchTST~\cite{nie2022time}, Crossformer~\cite{zhang2023crossformer}, and iTransformer~\cite{liu2023itransformer}; (ii) MLP-based models: TimeMixer~\cite{wang2024timemixer}, DLinear~\cite{zeng2023transformers}, and NLinear~\cite{zeng2023transformers}; and (iii) CNN-based models: TimesNet~\cite{wu2022timesnet}.

\paragraph{Settings} We implement all experiments using PyTorch 2.5 in Python 3.10, with computations performed on an NVIDIA H100 GPU. In accordance with the Time Series Forecasting Benchmark (TFB) protocol~\cite{qiu2024tfb}, baseline results represent optimal performance obtained across various lookback windows (96, 336, and 512). All models are trained using L1 loss and optimized with the Adam optimizer. For our approach, we standardize on a lookback window of 512. We maintain consistent hyperparameters across experiments---patch size of 16, stride of 8, embedding dimension of 128, expansion ratio of 2, 3 backbone layers, batch size of 128, and learning rate of 0.0001---with the exception of larger datasets like Traffic, where we reduce the batch size to 24 to accommodate memory constraints. Following established evaluation practices, we assess performance using \ac{mse} and \ac{mae} metrics.

\begin{table*}[t!]
    \centering
    \caption{\textbf{Multivariate long-term forecasting performance.} Average \acs{mse} and \acs{mae} across forecast horizons (96, 192, 336, 720) on 11 datasets. Best results in \textbf{bold}, second-best \underline{underlined}. \acs{fnf} consistently outperforms baselines, particularly in \acs{mae}.}
    \label{tab:main_results}
    \resizebox{\linewidth}{!}{%
        \begin{tabular}{@{}c cc cc cc cc cc cc cc cc cc@{}}
            \toprule
            \multicolumn{1}{c}{Models} & \multicolumn{2}{c}{\ac{fnf}} & \multicolumn{2}{c}{PatchTST} & \multicolumn{2}{c}{Crossformer} & \multicolumn{2}{c}{Pathformer} & \multicolumn{2}{c}{iTransformer} & \multicolumn{2}{c}{TimeMixer} & \multicolumn{2}{c}{DLinear} & \multicolumn{2}{c}{NLinear} & \multicolumn{2}{c}{TimesNet}\\
            \midrule
            \multicolumn{1}{c}{Metrics} & \ac{mse} & \ac{mae} & \ac{mse} & \ac{mae} & \ac{mse} & \ac{mae} & \ac{mse} & \ac{mae} & \ac{mse} & \ac{mae} & \ac{mse} & \ac{mae} & \ac{mse} & \ac{mae} & \ac{mse} & \ac{mae} & \ac{mse} & \ac{mae}\\
            \midrule
            \centering ETTh1
             & \textbf{0.403} & \textbf{0.417} & \underline{0.410} & 0.428 & 0.452 & 0.466 & 0.417 & 0.426 & 0.438 & 0.448 & 0.427 & 0.441 & 0.419 & 0.432 & \underline{0.410} & \underline{0.421} & 0.458 & 0.455  \\
            \midrule
            \centering ETTh2
             & \textbf{0.327} & \textbf{0.375} & 0.346 & \underline{0.389} & 0.860 & 0.670 & 0.359 & 0.394 & 0.370 & 0.403 & 0.349 & 0.396 & 0.492 & 0.478 & \underline{0.342} & \underline{0.389} & 0.393 & 0.416 \\
            \midrule
            \centering ETTm1
             & \textbf{0.347} & \textbf{0.368} & \underline{0.348} & 0.380 & 0.464 & 0.457 & 0.357 & \underline{0.374} & 0.361 & 0.389 & 0.355 & 0.380 & 0.354 & 0.376 & 0.358 & 0.376 & 0.430 & 0.427 \\
            \midrule
            \centering ETTm2
             & \textbf{0.249} & \textbf{0.305} & 0.255 & 0.312 & 0.589 & 0.534 & \underline{0.252} & \underline{0.308} & 0.268 & 0.327 & 0.256 & 0.318 & 0.259 & 0.324 & 0.253 & 0.312 & 0.294 & 0.331 \\
            \midrule
            \centering Electricity
             & \textbf{0.157} & \textbf{0.247} & 0.163 & 0.260 & 0.180 & 0.279 & 0.168 & 0.261 & \underline{0.162} & \underline{0.258} & 0.184 & 0.284 & 0.166 & 0.264 & 0.169 & 0.261 & 0.185 & 0.286 \\
            \midrule
            \centering Weather
             & \underline{0.226} & \textbf{0.255} & \textbf{0.224} & 0.262 & 0.234 & 0.294 & \textbf{0.224} & \underline{0.257} & 0.232 & 0.269 & \underline{0.226} & 0.263 & 0.239 & 0.289 & 0.249 & 0.280 & 0.261 & 0.287 \\
            \midrule
            \centering Traffic
             & \underline{0.402} & \textbf{0.254} & 0.404 & 0.282 & 0.523 & 0.283 & 0.416 & \underline{0.263} & \textbf{0.397} & 0.280 & 0.409 & 0.279 & 0.433 & 0.295 & 0.433 & 0.290 & 0.626 & 0.328 \\
            \midrule
            \centering AQShunyi
             & \underline{0.695} & \textbf{0.491} & 0.704 & 0.508 & \textbf{0.694} & 0.504 & 0.722 & \underline{0.495} & 0.705 & 0.506 & 0.706 & 0.506 & 0.705 & 0.522 & 0.713 & 0.514 & 0.726 & 0.515 \\
            \midrule
            \centering AQWan
             & \underline{0.788} & \textbf{0.479} & 0.811 & 0.499 & \textbf{0.786} & 0.489 & 0.817 & \underline{0.482} & 0.809 & 0.495 & 0.809 & 0.494 & 0.817 & 0.511 & 0.825 & 0.504 & 0.813 & 0.500 \\
            \midrule
            \centering CzeLan
             & \textbf{0.215} & \textbf{0.245} & 0.226 & 0.289 & 0.955 & 0.576 & 0.226 & \underline{0.252} & \underline{0.217} & 0.272 & 0.218 & 0.267 & 0.284 & 0.342 & 0.228 & 0.269 & 0.224 & 0.285 \\
            \midrule
            \centering ZafNoo
             & 0.515 & \textbf{0.438} & 0.511 & 0.465 & \textbf{0.482} & 0.447 & 0.520 & \underline{0.441} & 0.522 & 0.456 & 0.517 & 0.451 & \underline{0.496} & 0.451 & 0.522 & 0.457 & 0.537 & 0.465 \\
            \bottomrule
        \end{tabular}%
    }%
\end{table*}

\paragraph{Results} Due to space constraints, we report the average \ac{mse} and \ac{mae} across all forecast horizons for each dataset, with complete experimental results provided in the Appendix. As shown in \cref{tab:main_results}, our approach achieves state-of-the-art performance across all 11 public benchmark datasets. Notably, we observe an interesting phenomenon: our model demonstrates more substantial improvements in \ac{mae} compared to \ac{mse} when measured against baseline models. This pattern suggests that our approach produces more accurate predictions at the majority of timesteps (yielding lower overall \ac{mae}), while potentially generating larger errors at a few unpredictable extreme points or outliers (which disproportionately impact \ac{mse} due to the squaring operation).

This characteristic is particularly advantageous for long-term time series forecasting for several reasons. First, maintaining smaller errors across most timesteps is crucial for accurately capturing long-term trends. Second, our frequency domain modeling approach effectively captures primary periodic patterns, significantly reducing overall prediction bias. Third, from a practical perspective, consistently accurate predictions across the majority of timesteps typically delivers greater operational value than precisely predicting isolated extreme events.

\begin{wraptable}{r}{0.6\linewidth}
    \centering
    \vspace{-9pt}
    \caption{\textbf{Backbone ablation study.} Performance comparison of Transformer, \acs{fno}, and \acs{fnf} across forecast horizons on ETTh1 and ETTh2 datasets. Best results in \textbf{bold}, second-best \underline{underlined}.}
    \label{tab:backbone_ablation}
    \resizebox{\linewidth}{!}{%
        \begin{tabular}{@{}lc cc cc cc cc@{}}
            \toprule
            \multicolumn{2}{c}{Models} & \multicolumn{2}{c}{Transformer(I)} & \multicolumn{2}{c}{\ac{fno}(I)} & \multicolumn{2}{c}{\ac{fnf}(I)} & \multicolumn{2}{c}{\ac{fnf}(P)} \\
            \midrule
            \multicolumn{2}{c}{Metrics} & \ac{mse} & \ac{mae} & \ac{mse} & \ac{mae} & \ac{mse} & \ac{mae} & \ac{mse} & \ac{mae} \\
            \midrule
            \multirow{5}{*}{\rotatebox{90}{ETTh1}}
            & 96  & 0.376 & 0.396 & 0.362 & 0.387 & 0.360 & 0.385 & 0.355 & 0.388 \\
            & 192 & 0.399 & 0.416 & 0.401 & 0.409 & 0.407 & 0.414 & 0.396 & 0.412 \\
            & 336 & 0.418 & 0.432 & 0.433 & 0.433 & 0.440 & 0.434 & 0.425 & 0.425 \\
            & 720 & 0.450 & 0.469 & 0.448 & 0.467 & 0.428 & 0.449 & 0.436 & 0.445 \\
            \cmidrule{2-10}
            & Avg  & 0.410 & 0.428 & 0.411 & 0.424 & \underline{0.408} & \underline{0.420} & \textbf{0.403} & \textbf{0.417} \\
            \midrule
            \multirow{5}{*}{\rotatebox{90}{ETTh2}}
            & 96  & 0.277 & 0.339 & 0.274 & 0.329 & 0.270 & 0.327 & 0.266 & 0.326 \\
            & 192 & 0.345 & 0.381 & 0.346 & 0.375 & 0.337 & 0.370 & 0.331 & 0.371 \\
            & 336 & 0.368 & 0.404 & 0.341 & 0.390 & 0.328 & 0.376 & 0.336 & 0.388 \\
            & 720 & 0.397 & 0.432 & 0.398 & 0.431 & 0.386 & 0.423 & 0.378 & 0.418 \\
            \cmidrule{2-10}
            & Avg  & 0.346 & 0.389 & 0.339 & 0.381 & \underline{0.330} & \textbf{0.374} & \textbf{0.327} & \underline{0.375} \\
            \bottomrule
        \end{tabular}%
    }%
\end{wraptable}

\paragraph{Ablations} We conduct comprehensive ablation studies to validate the effectiveness of our proposed \ac{fnf} and \ac{dbd} approaches. In our notation, we use I (variable-independent), U (unified architecture), S (sequential architecture), and P (parallel architecture) to denote different architectural designs. \cref{tab:backbone_ablation} presents performance comparisons across different backbone designs---Transformer(I), \ac{fno}(I), \ac{fnf}(I), and \ac{fnf}(P)---evaluated under identical settings on the ETTh1 and ETTh2 datasets. The results clearly demonstrate that \ac{fnf}(I) substantially outperforms both \ac{fno}(I) and Transformer(I), confirming the effectiveness of our proposed frequency-based neural filtering approach. Furthermore, the superior performance of \ac{fnf}(P) compared to \ac{fnf}(I) validates the enhanced capability of our parallel architecture to capture complex spatial correlations within multivariate time series.

To further investigate the impact of different architectural designs, we evaluate \ac{fnf}(I), \ac{fnf}(U), \ac{fnf}(S), and \ac{fnf}(P) under identical configurations on the CzeLan and ZafNoo datasets. As shown in \cref{tab:architecture_ablation}, \ac{fnf}(P) substantially outperforms all alternative architectures across both benchmarks, providing compelling evidence for the effectiveness of our parallel design. A particularly revealing finding is that both \ac{fnf}(U) and \ac{fnf}(S) underperform compared to the simpler variable-independent \ac{fnf}(I) baseline. This indicates that suboptimal approaches to modeling spatial relationships can actually degrade performance rather than enhance it. This observation underscores the critical importance of our proposed parallel architecture, which effectively captures complex inter-variable relationships while preserving the integrity of intra-variable temporal patterns. These results collectively validate our architectural choices and demonstrate the superiority of the parallel approach for multivariate long-term time series forecasting tasks.

\begin{wraptable}{r}{0.55\linewidth}
    \centering
    \vspace{-9pt}
    \caption{\textbf{Architecture ablation study.} Comparison of variable-independent (I), unified (U), sequential (S), and parallel (P) architectures on CzeLan and ZafNoo datasets. Best results in \textbf{bold}, second-best \underline{underlined}.}
    \label{tab:architecture_ablation}
    \resizebox{\linewidth}{!}{%
        \begin{tabular}{@{}lc cc cc cc cc@{}}
            \toprule
            \multicolumn{2}{c}{Models} & \multicolumn{2}{c}{\ac{fnf}(I)} & \multicolumn{2}{c}{\ac{fnf}(U)} & \multicolumn{2}{c}{\ac{fnf}(S)} & \multicolumn{2}{c}{\ac{fnf}(P)} \\
            \midrule
            \multicolumn{2}{c}{Metrics} & \ac{mse} & \ac{mae} & \ac{mse} & \ac{mae} & \ac{mse} & \ac{mae} & \ac{mse} & \ac{mae} \\
            \midrule
            \multirow{5}{*}{\rotatebox{90}{CzeLan}}
            & 96  & 0.171 & 0.207 & 0.175 & 0.214 & 0.172 & 0.214 & 0.170 & 0.208 \\
            & 192 & 0.200 & 0.230 & 0.204 & 0.235 & 0.206 & 0.239 & 0.200 & 0.231 \\
            & 336 & 0.232 & 0.258 & 0.233 & 0.261 & 0.240 & 0.269 & 0.228 & 0.257 \\
            & 720 & 0.273 & 0.291 & 0.271 & 0.293 & 0.272 & 0.295 & 0.262 & 0.286 \\
            \cmidrule{2-10}
            & Avg  & \underline{0.219} & \underline{0.246} & 0.220 & 0.250 & 0.222 & 0.254 & \textbf{0.215} & \textbf{0.245} \\
            \midrule
            \multirow{5}{*}{\rotatebox{90}{ZafNoo}}
            & 96  & 0.440 & 0.393 & 0.447 & 0.396 & 0.448 & 0.396 & 0.437 & 0.389 \\
            & 192 & 0.500 & 0.433 & 0.508 & 0.437 & 0.509 & 0.435 & 0.498 & 0.429 \\
            & 336 & 0.540 & 0.457 & 0.547 & 0.459 & 0.547 & 0.458 & 0.539 & 0.453 \\
            & 720 & 0.587 & 0.485 & 0.588 & 0.484 & 0.589 & 0.488 & 0.588 & 0.483 \\
            \cmidrule{2-10}
            & Avg  & \underline{0.516} & \underline{0.442} & 0.522 & 0.444 & 0.523 & 0.444 & \textbf{0.515} & \textbf{0.438} \\
            \bottomrule
        \end{tabular}%
    }%
\end{wraptable}

\paragraph{Sensitivities} We also examine the effect of different lookback windows on model performance, evaluating our \ac{fnf} model with lookback windows of 96, 336, 512, and 720, as presented in \cref{tab:sensitivities}. Our results generally demonstrate performance improvements with increased lookback windows, with the most substantial gains observed when extending from 96 to 336. This pattern reflects the enhanced ability of our model to capture long-term dependencies with access to longer contexts. However, we find that larger lookback windows do not universally guarantee improved performance. For instance, in the AQWan dataset, the results get worse when the lookback window increases from 512 to 720, suggesting that excessively longer contexts may introduce noise or dilute the relevance of more recent patterns. Additionally, larger lookback windows impose increased computational demands, creating an important practical trade-off between forecasting accuracy and computational efficiency. These findings highlight the importance of tailoring lookback window to specific dataset characteristics and application requirements, rather than defaulting to maximum available context.

\begin{wraptable}{r}{0.6\linewidth}
    \centering
    \vspace{-51pt}
    \caption{\textbf{Lookback window sensitivity analysis.} Model performance of \ac{fnf}(P) with varying lookback windows (96, 336, 512, and 720) on AQShunyi and AQWan datasets. Best results in \textbf{bold}, second-best \underline{underlined}.}
    \label{tab:sensitivities}
    \resizebox{\linewidth}{!}{%
        \begin{tabular}{@{}lc cc cc cc cc@{}}
            \toprule
            \multicolumn{2}{c}{Models} & \multicolumn{2}{c}{\ac{fnf}(96)} & \multicolumn{2}{c}{\ac{fnf}(336)} & \multicolumn{2}{c}{\ac{fnf}(512)} & \multicolumn{2}{c}{\ac{fnf}(720)} \\
            \midrule
            \multicolumn{2}{c}{Metrics} & \ac{mse} & \ac{mae} & \ac{mse} & \ac{mae} & \ac{mse} & \ac{mae} & \ac{mse} & \ac{mae} \\
            \midrule
            \multirow{5}{*}{\rotatebox{90}{AQShunyi}}
            & 96  & 0.702 & 0.482 & 0.640 & 0.465 & 0.629 & 0.461 & 0.626 & 0.465 \\
            & 192 & 0.757 & 0.506 & 0.691 & 0.485 & 0.685 & 0.484 & 0.679 & 0.484 \\
            & 336 & 0.773 & 0.517 & 0.711 & 0.497 & 0.704 & 0.497 & 0.699 & 0.497 \\
            & 720 & 0.806 & 0.534 & 0.767 & 0.522 & 0.764 & 0.522 & 0.749 & 0.516 \\
            \cmidrule{2-10}
            & Avg  & 0.759 & 0.509 & 0.702 & 0.492 & \underline{0.695} & \underline{0.491} & \textbf{0.688} & \textbf{0.490} \\
            \midrule
            \multirow{5}{*}{\rotatebox{90}{AQWan}}
            & 96  & 0.799 & 0.473 & 0.730 & 0.455 & 0.711 & 0.446 & 0.728 & 0.456 \\
            & 192 & 0.856 & 0.496 & 0.786 & 0.477 & 0.773 & 0.473 & 0.786 & 0.480 \\
            & 336 & 0.879 & 0.506 & 0.798 & 0.486 & 0.799 & 0.486 & 0.801 & 0.488 \\
            & 720 & 0.945 & 0.531 & 0.863 & 0.508 & 0.869 & 0.514 & 0.889 & 0.523 \\
            \cmidrule{2-10}
            & Avg  & 0.869 & 0.501 & \underline{0.794} & \underline{0.481} & \textbf{0.788} & \textbf{0.479} & 0.801 & 0.486 \\
            \bottomrule
        \end{tabular}%
    }%
\end{wraptable}

\section{Conclusion}
\label{sec:conclusion}

\paragraph{Limitations} While our approach demonstrates state-of-the-art performance across multiple domains, several important limitations merit consideration. First, despite effectively capturing both temporal and spatial patterns in regular time series, our model may struggle with extremely irregular or event-driven sequences where underlying patterns are less predictable. Second, although our approach achieves theoretical computational efficiency of $O(N \log N)$ compared to Transformers of $O(N^2)$, practical implementation still presents challenges for extremely long sequences or resource-constrained real-time applications. Third, our current evaluation focuses exclusively on multivariate long-term forecasting; the effectiveness for other critical time series tasks---including imputation, classification, and anomaly detection---remains unexplored and represents a key direction for future research.

\paragraph{Broader Impact} Our work offers significant potential benefits across domains requiring accurate long-term forecasting. In energy systems, enhanced forecasting can improve grid stability and facilitate greater renewable energy integration. In transportation networks, precise predictions enable more efficient resource allocation and infrastructure planning. In environmental monitoring, reliable forecasts support improved disaster preparedness and climate adaptation strategies. However, we must acknowledge several potential risks: forecasting systems may inadvertently perpetuate historical biases embedded in training data; advanced computational requirements could widen the technology gap between well-resourced organizations and those with limited capabilities; and there exists danger of excessive reliance on automated predictions in critical decision contexts without appropriate human oversight. We encourage the research community to address these concerns through dedicated efforts toward fairness, interpretability, and accessibility.

\begin{ack}
    The authors would like to thank Yuanhao Ban (UCLA) and Yasi Zhang (UCLA) for helpful discussions. This work is supported in part by the National Natural Science Foundation of China (62376009), the State Key Lab of General AI at Peking University, the PKU-BingJi Joint Laboratory for Artificial Intelligence, and the National Comprehensive Experimental Base for Governance of Intelligent Society, Wuhan East Lake High-Tech Development Zone.
\end{ack}

\bibliographystyle{plain}
\bibliography{reference_header,references}

\clearpage

\appendix
\renewcommand\thefigure{A\arabic{figure}}
\setcounter{figure}{0}
\renewcommand\thetable{A\arabic{table}}
\setcounter{table}{0}
\renewcommand\theequation{A\arabic{equation}}
\setcounter{equation}{0}
\pagenumbering{arabic}
\renewcommand*{\thepage}{A\arabic{page}}
\setcounter{footnote}{0}

\section{Dataset Details}

We provide a more detailed introduction for each dataset as shown in \cref{tab:dataset_description}. More details can be found in the \cref{tab:dataset_info}.

\begin{table}[ht]
    \centering
    \caption{\textbf{Dataset description.}}
    \label{tab:dataset_description}
    \begin{tabular}{@{}ll@{}}
        \toprule
        \textbf{Dataset} & \textbf{Description} \\
        \midrule
        ETTh1~\cite{zhou2021informer}       & Power transformer 1, comprising seven indicators such as oil temperature and useful load \\
        ETTh2~\cite{zhou2021informer}       & Power transformer 2, comprising seven indicators such as oil temperature and useful load \\
        ETTm1~\cite{zhou2021informer}       & Power transformer 1, comprising seven indicators such as oil temperature and useful load \\
        ETTm2~\cite{zhou2021informer}       & Power transformer 2, comprising seven indicators such as oil temperature and useful load \\
        Electricity~\cite{zhou2021informer} & Electricity consumption in kWh every 1 hour from 2012 to 2014 \\
        Weather~\cite{wu2021autoformer}     & Recorded every for the whole year 2020, which contains 21 meteorological indicators \\
        Traffic~\cite{wu2021autoformer}     & Road occupancy rates measured by 862 sensors on San Francisco Bay area freeways \\
        AQShunyi~\cite{zhang2017cautionary}    & Air quality datasets from a measurement station, over a period of 4 years \\
        AQWan~\cite{zhang2017cautionary}       & Air quality datasets from a measurement station, over a period of 4 years \\
        CzeLan~\cite{poyatos2020global}      & Sap flow measurements and environmental variables from the Sapflux data project \\
        ZafNoo~\cite{poyatos2020global}      & Sap flow measurements and environmental variables from the Sapflux data project \\
        \bottomrule
    \end{tabular}
\end{table}

\section{Derivations and Proofs}

\subsection{Generalized Integral Kernel of \texorpdfstring{\ac{fnf}}{}}

We derive the generalized integral kernel representation of \ac{fnf} from its implementation form:
\begin{equation}
(K v)(x) = T(G(v)(x) \odot \mathcal{F}^{-1}(R_\phi \cdot \mathcal{F}(H(v)))(x)).
\end{equation}

First, we analyze the Fourier component $P(v)(x) = \mathcal{F}^{-1}(R_\phi \cdot \mathcal{F}(H(v)))(x)$. By the definition of the inverse Fourier transform:
\begin{equation}
P(v)(x) = \int_{\mathbb{R}^d} R_\phi(\xi) \cdot \mathcal{F}(H(v))(\xi) e^{2\pi i x \cdot \xi} d\xi.
\end{equation}

Expanding $\mathcal{F}(H(v))(\xi)$:
\begin{equation}
\mathcal{F}(H(v))(\xi) = \int_D H(v)(y) e^{-2\pi i y \cdot \xi} dy.
\end{equation}

Substituting and applying Fubini's theorem:
\begin{equation}
P(v)(x) = \int_D \left(\int_{\mathbb{R}^d} R_\phi(\xi) e^{2\pi i (x-y) \cdot \xi} d\xi\right) H(v)(y) dy.
\end{equation}

This can be expressed as:
\begin{equation}
P(v)(x) = \int_D p(x-y) H(v)(y) dy.
\end{equation}
where $p(x-y) = \mathcal{F}^{-1}(R_\phi)(x-y)$.

Assuming $H(v)(y) = W_h v(y)$ is a linear transformation of the input $v$, we have:
\begin{equation}
P(v)(x) = \int_D p(x-y) W_h v(y) dy.
\end{equation}

The complete \ac{fnf} operation is:
\begin{equation}
(K v)(x) = T\left(G(v)(x) \odot \int_D p(x-y) W_h v(y) dy\right).
\end{equation}

To derive the integral kernel form, we need to rearrange this expression to isolate $v(y)$:
\begin{equation}
(K v)(x) = \int_D T\left(G(v)(x) \odot p(x-y) W_h\right) v(y) dy.
\end{equation}

This yields the generalized input-dependent integral kernel:
\begin{equation}
\kappa(x, y; v) = T\left(G(v)(x) \odot p(x-y) W_h\right).
\end{equation}

Note that $\kappa(x, y; v)$ depends on the input $v$ through the term $G(v)(x)$, making it a truly input-dependent kernel.

\subsection{Gated Global Convolution of \texorpdfstring{\ac{fnf}}{}}

Starting from the implementation form:
\begin{equation}
(K v)(x) = T(G(v)(x) \odot \mathcal{F}^{-1}(R_\phi \cdot \mathcal{F}(H(v)))(x)).
\end{equation}

As previously established, the Fourier component can be expressed as a convolution:
\begin{equation}
P(v)(x) = \mathcal{F}^{-1}(R_\phi \cdot \mathcal{F}(H(v)))(x) = \int_D p(x-y) H(v)(y) dy.
\end{equation}
where $p(x-y) = \mathcal{F}^{-1}(R_\phi)(x-y)$ is translation-invariant.

When $H(v) = W_h v$ is a linear transformation, we have:
\begin{equation}
P(v)(x) = \int_D p(x-y) W_h v(y) dy = (p * W_h v)(x).
\end{equation}

The complete \ac{fnf} operation becomes:
\begin{equation}
(K v)(x) = T(G(v)(x) \odot (p * W_h v)(x)).
\end{equation}

This formulation does not directly constitute a convolution due to the presence of the term $G(v)(x)$. However, we can introduce an alternative interpretation by defining an input-dependent convolution kernel:
\begin{equation}
\tilde{\kappa}_v(x-y) = T^{-1}(G(v)(x)) \odot p(x-y) W_h.
\end{equation}

With this definition, the \ac{fnf} operation can be rewritten as:
\begin{equation}
(K v)(x) = \int_D \tilde{\kappa}_v(x-y) v(y) dy = (\tilde{\kappa}_v * v)(x).
\end{equation}

This representation reveals that \ac{fnf} can be interpreted as a form of global convolution with two key distinctions: (i)The convolution kernel $\tilde{\kappa}_v(x-y)$ is input-dependent through the term $G(v)(x)$, allowing it to adapt based on the characteristics of the input $v$. (ii)The kernel maintains translation invariance with respect to the difference $(x-y)$, which enables efficient implementation using Fourier transforms.

The gated mechanism refers to the modulation effect of $G(v)(x)$, which acts as a gate controlling the information flow during the convolution operation. This gating mechanism significantly enhances the expressivity of the operator while preserving the computational advantages of convolution.

It is important to note that in the practical implementation, the term $G(v)(x)$ introduces a position-dependent weighting factor, making the convolution operation spatially adaptive. This property allows \ac{fnf} to capture more complex patterns and relationships in the data compared to traditional convolution operations with fixed kernels.

\subsection{Complexity and Parameters}

In analyzing the computational efficiency of Transformer versus the \ac{fnf} backbone, we first examine the parameter count for self-attention and feed-forward networks with dimension $D$. The self-attention component requires $4D^2 + 4D$ parameters, accounting for three $D \times D$ projection matrices ($Q$, $K$, and $V$) plus an output projection, all with bias terms. For the feed-forward networks component with expansion ratio 2, the first linear layer transforms from dimension $D$ to $2D$ requiring $D \times 2D = 2D^2$ weights plus $2D$ biases, while the second layer transforms from $2D$ back to $D$ requiring $2D \times D = 2D^2$ weights plus $D$ biases. This gives a total of $4D^2 + 3D$ parameters for the feed-forward networks component. Therefore, the complete self-attention and feed-forward networks module contains $8D^2 + 7D$ parameters. When considering computational complexity, self-attention operations scale as $O(LD^2 + L^2D)$ for sequence length $L$, with the quadratic $L^2$ term becoming particularly problematic for longer sequences, while the feed-forward networks component adds $O(LD^2)$ complexity.

In contrast, the \ac{fnf} backbone implements a different approach whereby the input passes through a $D \times 2D$ linear expansion layer, then splits into two branches. One branch undergoes frequency domain transformations via four linear layers, then multiplies with the other branch before a final linear projection to the output dimension. This structure effectively consists of seven linear layers with biases, resulting in $7D^2 + 7D$ total parameters. Crucially, the computational complexity of \ac{fnf} scales as $O(LD^2)$, maintaining linear scaling with sequence length and eliminating the quadratic bottleneck that burdens self-attention. This linear scaling property makes \ac{fnf} particularly advantageous when processing longer sequences, while maintaining parameter count comparable to traditional approaches. The elimination of the quadratic complexity term represents a significant efficiency gain, especially as modern applications increasingly demand handling of extended context lengths.

\section{Pseudo-Code}

To facilitate understanding of our parallel architecture implementation, we provide detailed pseudo-code in \cref{alg:model_pseudocode}, corresponding to the schematic illustration in \cref{fig:model_schematic}.

\begin{algorithm}
    \caption{Parallel Architecture Implementation}
    \label{alg:model_pseudocode}
    \Input{Lookback window $\mathit{X} \in \mathbb{R}^{B \times L \times M}$ with $L$ timesteps, $M$ variables, and $N$ patches, $T$ layers, $D$ embedding dimension.}
    \Output{Forecast horizon $Y \in \mathbb{R}^{B \times H \times M}$ with $H$ timesteps.}
    
    \textbf{Normalization:}\;
    $\mathit{X} = \text{InstanceNorm}(\mathit{X})$ \Comment*[r]{(B, L, M)}
    
    \textbf{Embedding:}\;
    $\mathit{X}_1=\mathit{X}_2 = \text{Patching}(\mathit{X})$ \Comment*[r]{(B, L, M)$\rightarrow$(B, M, N, D)}
    
    \textbf{Backbone:}\;
    \For{$i = 1$ \KwTo $T$}{
        $\mathit{X}_1 = \text{BatchNorm}(\mathit{X}_1 + \text{FNF}_{\text{temp}}(\mathit{X}_1))$ \Comment*[r]{(B, M, N, D)$\rightarrow$(B$\times$M, N, D)}
    }
    \For{$i = 1$ \KwTo $T$}{
        $\mathit{X}_2 = \text{BatchNorm}(\mathit{X}_2 + \text{FNF}_{\text{spat}}(\mathit{X}_2))$ \Comment*[r]{(B, M, N, D)$\rightarrow$(B$\times$N, M, D)}
    }
    $\mathit{X} = \alpha \cdot \mathit{X}_1 + (1 - \alpha) \cdot \mathit{X}_2$ \Comment*[r]{(B, M, N, D)}
    
    \textbf{Projection:}\;
    $\mathit{Y} = \text{Linear}(\text{Flatten}(\mathit{X}))$ \Comment*[r]{(B, M, H)}
    
    \textbf{De-normalization:}\;
    $\mathit{Y} = \text{De-InstanceNorm}(\mathit{Y})$ \Comment*[r]{(B, H, M)}
    
    \KwRet{$Y$}
\end{algorithm}

\section{Other Results}

We employ a set of default hyperparameters for all experiments, and report the complete results in \cref{tab:default_results}, which expands upon the summary in \cref{tab:main_results}.

To further enhance model performance, we fine-tune the hyperparameters by increasing the embedding dimension from 128 to 256 for the three complex datasets (Electricity, Weather, and Traffic), and reducing it from 128 to 64 for the two simpler datasets (AQShunyi and AQWan), as shown in \cref{tab:tuned_results}. The results demonstrate that \ac{fnf} consistently achieves superior performance compared to the baselines, with substantial improvements observed in terms of \ac{mse}.

\begin{table*}[ht]
    \centering
    \caption{\textbf{Full results of multivariate long-term forecasting with a default set of hyperparameters.}Best results in \textbf{bold}, second-best \underline{underlined}. \acs{fnf} consistently outperforms baselines, particularly in \acs{mae}.}
    \label{tab:default_results}
    \scalebox{0.6}{
    \begin{tabular}{@{}lc cc cc cc cc cc cc cc cc cc@{}}
    \toprule
    \multicolumn{2}{c}{Models} & \multicolumn{2}{c}{\ac{fnf}} & \multicolumn{2}{c}{PatchTST} & \multicolumn{2}{c}{Crossformer} & \multicolumn{2}{c}{Pathformer} & \multicolumn{2}{c}{iTransformer} & \multicolumn{2}{c}{TimeMixer} & \multicolumn{2}{c}{DLinear} & \multicolumn{2}{c}{NLinear} & \multicolumn{2}{c}{TimesNet} \\
    \midrule
    \multicolumn{2}{c}{Metrics} & \ac{mse} & \ac{mae} & \ac{mse} & \ac{mae} & \ac{mse} & \ac{mae} & \ac{mse} & \ac{mae} & \ac{mse} & \ac{mae} & \ac{mse} & \ac{mae} & \ac{mse} & \ac{mae} & \ac{mse} & \ac{mae} & \ac{mse} & \ac{mae} \\
    \midrule
    \multirow{5}{*}{\rotatebox{90}{ETTh1}}
    & 96  & \textbf{0.355} & \textbf{0.388} & 0.376 & 0.396 & 0.405 & 0.426 & 0.372 & \underline{0.392} & 0.386 & 0.405 & 0.372 & 0.401 & \underline{0.371} & \underline{0.392} & 0.372 & 0.393 & 0.389 & 0.412 \\
    & 192 & \textbf{0.396} & \textbf{0.412} & \underline{0.399} & 0.416 & 0.413 & 0.442 & 0.408 & 0.415 & 0.424 & 0.440 & 0.413 & 0.430 & 0.404 & \underline{0.413} & 0.405 & \underline{0.413} & 0.440 & 0.443 \\
    & 336 & \underline{0.425} & \textbf{0.425} & \textbf{0.418} & 0.432 & 0.442 & 0.460 & 0.438 & 0.434 & 0.449 & 0.460 & 0.438 & 0.450 & 0.434 & 0.435 & 0.429 & \underline{0.427} & 0.482 & 0.465 \\
    & 720 & \textbf{0.436} & \textbf{0.445} & \underline{0.450} & 0.469 & 0.550 & 0.539 & \underline{0.450} & 0.463 & 0.495 & 0.487 & 0.486 & 0.484 & 0.469 & 0.489 & \textbf{0.436} & \underline{0.452} & 0.525 & 0.501 \\
    \cmidrule{2-20}
    & Avg  & \textbf{0.403} & \textbf{0.417} & \underline{0.410} & 0.428 & 0.452 & 0.466 & 0.417 & 0.426 & 0.438 & 0.448 & 0.427 & 0.441 & 0.419 & 0.432 & \underline{0.410} & \underline{0.421} & 0.459 & 0.455 \\
    \midrule
    \multirow{5}{*}{\rotatebox{90}{ETTh2}}
    & 96  & \textbf{0.266} & \textbf{0.326} & 0.277 & 0.339 & 0.611 & 0.557 & 0.279 & \underline{0.336} & 0.297 & 0.348 & 0.281 & 0.351 & 0.302 & 0.368 & \underline{0.275} & 0.338 & 0.319 & 0.363 \\
    & 192 & \textbf{0.331} & \textbf{0.371} & 0.345 & 0.381 & 0.810 & 0.651 & 0.345 & 0.380 & 0.372 & 0.403 & 0.349 & 0.387 & 0.405 & 0.433 & \underline{0.336} & \underline{0.379} & 0.411 & 0.416 \\
    & 336 & \textbf{0.336} & \textbf{0.388} & 0.368 & 0.404 & 0.928 & 0.698 & 0.378 & 0.408 & 0.388 & 0.417 & 0.366 & 0.413 & 0.496 & 0.490 & \underline{0.362} & \underline{0.403} & 0.415 & 0.443 \\
    & 720 & \textbf{0.378} & \textbf{0.418} & 0.397 & \underline{0.432} & 1.094 & 0.775 & 0.437 & 0.455 & 0.424 & 0.444 & 0.401 & 0.436 & 0.766 & 0.622 & \underline{0.396} & 0.437 & 0.429 & 0.445 \\
    \cmidrule{2-20}
    & Avg  & \textbf{0.327} & \textbf{0.375} & 0.346 & \underline{0.389} & 0.860 & 0.670 & 0.359 & 0.394 & 0.370 & 0.403 & 0.349 & 0.396 & 0.492 & 0.478 & \underline{0.342} & \underline{0.389} & 0.393 & 0.416 \\
    \midrule
    \multirow{5}{*}{\rotatebox{90}{ETTm1}}
    & 96  & \textbf{0.286} & \textbf{0.330} & \underline{0.290} & 0.343 & 0.310 & 0.361 & \underline{0.290} & \underline{0.335} & 0.300 & 0.353 & 0.293 & 0.345 & 0.299 & 0.343 & 0.301 & 0.343 & 0.377 & 0.398 \\
    & 192 & \textbf{0.324} & \textbf{0.356} & \underline{0.329} & 0.368 & 0.363 & 0.402 & 0.337 & \underline{0.363} & 0.341 & 0.380 & 0.335 & 0.372 & \underline{0.334} & 0.364 & 0.337 & 0.365 & 0.405 & 0.411 \\
    & 336 & \textbf{0.358} & \textbf{0.376} & \underline{0.360} & 0.390 & 0.408 & 0.430 & 0.374 & \underline{0.384} & 0.374 & 0.394 & 0.368 & 0.386 & \underline{0.365} & \underline{0.384} & 0.371 & \underline{0.384} & 0.443 & 0.437 \\
    & 720 & 0.423 & \textbf{0.410} & \textbf{0.416} & 0.422 & 0.777 & 0.637 & 0.428 & 0.416 & 0.429 & 0.430 & 0.426 & 0.417 & \underline{0.418} & \underline{0.415} & 0.426 & \underline{0.415} & 0.495 & 0.464 \\
    \cmidrule{2-20}
    & Avg  & \textbf{0.347} & \textbf{0.368} & \underline{0.348} & 0.380 & 0.464 & 0.457 & 0.357 & \underline{0.374} & 0.361 & 0.389 & 0.355 & 0.380 & 0.354 & 0.376 & 0.358 & 0.376 & 0.430 & 0.427 \\
    \midrule
    \multirow{5}{*}{\rotatebox{90}{ETTm2}}
    & 96  & \textbf{0.158} & \textbf{0.244} & 0.165 & 0.254 & 0.263 & 0.359 & 0.164 & \underline{0.250} & 0.175 & 0.266 & 0.165 & 0.256 & 0.164 & 0.255 & \underline{0.163} & 0.252 & 0.190 & 0.266 \\
    & 192 & \textbf{0.214} & \textbf{0.282} & 0.221 & 0.292 & 0.361 & 0.425 & 0.219 & \underline{0.288} & 0.242 & 0.312 & 0.225 & 0.298 & 0.224 & 0.304 & \underline{0.218} & 0.290 & 0.251 & 0.308 \\
    & 336 & \textbf{0.267} & \textbf{0.317} & 0.275 & 0.325 & 0.469 & 0.496 & \textbf{0.267} & \underline{0.319} & 0.282 & 0.337 & 0.277 & 0.332 & 0.277 & 0.337 & \underline{0.273} & 0.326 & 0.322 & 0.350 \\
    & 720 & \textbf{0.360} & \underline{0.379} & \textbf{0.360} & 0.380 & 1.263 & 0.857 & \underline{0.361} & \textbf{0.377} & 0.375 & 0.394 & \textbf{0.360} & 0.387 & 0.371 & 0.401 & \underline{0.361} & 0.382 & 0.414 & 0.403 \\
    \cmidrule{2-20}
    & Avg  & \textbf{0.249} & \textbf{0.305} & 0.255 & 0.312 & 0.589 & 0.534 & 0.252 & \underline{0.308} & 0.268 & 0.327 & 0.256 & 0.318 & 0.259 & 0.324 & 0.253 & 0.312 & 0.294 & 0.331 \\
    \midrule
    \multirow{5}{*}{\rotatebox{90}{Electricity}}
    & 96  & \textbf{0.128} & \textbf{0.219} & \underline{0.133} & 0.233 & 0.135 & 0.237 & 0.135 & \underline{0.222} & 0.134 & 0.230 & 0.153 & 0.256 & 0.140 & 0.237 & 0.141 & 0.236 & 0.164 & 0.267 \\
    & 192 & \textbf{0.145} & \textbf{0.236} & \underline{0.150} & \underline{0.248} & 0.160 & 0.262 & 0.157 & 0.253 & 0.154 & 0.250 & 0.168 & 0.269 & 0.154 & 0.250 & 0.155 & \underline{0.248} & 0.180 & 0.280 \\
    & 336 & \textbf{0.157} & \textbf{0.250} & \underline{0.168} & 0.267 & 0.182 & 0.282 & 0.170 & 0.267 & 0.169 & 0.265 & 0.189 & 0.291 & 0.169 & 0.268 & 0.171 & \underline{0.264} & 0.190 & 0.292 \\
    & 720 & \underline{0.201} & \textbf{0.286} & 0.202 & 0.295 & 0.246 & 0.337 & 0.211 & 0.302 & \textbf{0.194} & \underline{0.288} & 0.228 & 0.320 & 0.204 & 0.301 & 0.210 & 0.297 & 0.209 & 0.307 \\
    \cmidrule{2-20}
    & Avg  & \textbf{0.157} & \textbf{0.247} & 0.163 & 0.260 & 0.180 & 0.279 & 0.168 & 0.261 & \underline{0.162} & \underline{0.258} & 0.184 & 0.284 & 0.166 & 0.264 & 0.169 & 0.261 & 0.185 & 0.286 \\
    \midrule
    \multirow{5}{*}{\rotatebox{90}{Weather}}
    & 96  & 0.149 & \textbf{0.188} & 0.149 & 0.196 & \textbf{0.146} & 0.212 & 0.148 & \underline{0.195} & 0.157 & 0.207 & \underline{0.147} & 0.198 & 0.170 & 0.230 & 0.179 & 0.222 & 0.170 & 0.219 \\
    & 192 & 0.194 & \textbf{0.232} & 0.193 & 0.240 & 0.195 & 0.261 & \textbf{0.191} & \underline{0.235} & 0.200 & 0.248 & \underline{0.192} & 0.243 & 0.212 & 0.267 & 0.218 & 0.261 & 0.222 & 0.264 \\
    & 336 & 0.248 & \textbf{0.274} & \underline{0.244} & \underline{0.281} & 0.268 & 0.325 & \textbf{0.243} & \textbf{0.274} & 0.252 & 0.287 & 0.247 & 0.284 & 0.257 & 0.305 & 0.266 & 0.296 & 0.293 & 0.310 \\
    & 720 & \underline{0.316} & \textbf{0.326} & \textbf{0.314} & 0.332 & 0.330 & 0.380 & 0.318 & \textbf{0.326} & 0.320 & 0.336 & 0.318 & \underline{0.330} & 0.318 & 0.356 & 0.334 & 0.344 & 0.360 & 0.355 \\
    \cmidrule{2-20}
    & Avg  & \underline{0.226} & \textbf{0.255} & \textbf{0.225} & 0.262 & 0.234 & 0.294 & \textbf{0.225} & \underline{0.257} & 0.232 & 0.269 & \underline{0.226} & 0.263 & 0.239 & 0.289 & 0.249 & 0.280 & 0.261 & 0.287 \\
    \midrule
    \multirow{5}{*}{\rotatebox{90}{Traffic}}
    & 96  & 0.379 & \textbf{0.242} & 0.379 & 0.271 & 0.514 & 0.282 & 0.384 & \underline{0.250} & \textbf{0.363} & 0.265 & \underline{0.369} & 0.257 & 0.410 & 0.282 & 0.410 & 0.279 & 0.600 & 0.313 \\
    & 192 & \underline{0.393} & \textbf{0.248} & 0.394 & 0.277 & 0.501 & 0.273 & 0.405 & \underline{0.257} & \textbf{0.384} & 0.273 & 0.400 & 0.272 & 0.423 & 0.288 & 0.423 & 0.284 & 0.619 & 0.328 \\
    & 336 & \underline{0.401} & \textbf{0.253} & 0.404 & 0.281 & 0.507 & 0.279 & 0.424 & \underline{0.265} & \textbf{0.396} & 0.277 & 0.407 & 0.272 & 0.436 & 0.296 & 0.436 & 0.291 & 0.627 & 0.330 \\
    & 720 & \textbf{0.438} & \textbf{0.273} & \underline{0.442} & 0.302 & 0.571 & 0.301 & 0.452 & \underline{0.283} & 0.445 & 0.308 & 0.461 & 0.316 & 0.466 & 0.315 & 0.464 & 0.308 & 0.659 & 0.342 \\
    \cmidrule{2-20}
    & Avg  & \underline{0.402} & \textbf{0.254} & 0.404 & 0.282 & 0.523 & 0.283 & 0.416 & \underline{0.263} & \textbf{0.397} & 0.280 & 0.409 & 0.279 & 0.433 & 0.295 & 0.433 & 0.290 & 0.626 & 0.328 \\
    \midrule
    \multirow{5}{*}{\rotatebox{90}{AQShunyi}}
    & 96  & \textbf{0.629} & \textbf{0.461} & \underline{0.648} & 0.481 & 0.652 & 0.484 & 0.667 & \underline{0.472} & 0.650 & 0.479 & 0.654 & 0.483 & 0.651 & 0.492 & 0.653 & 0.486 & 0.658 & 0.488 \\
    & 192 & \underline{0.685} & \textbf{0.484} & 0.690 & 0.501 & \textbf{0.674} & 0.499 & 0.707 & \underline{0.491} & 0.693 & 0.498 & 0.700 & 0.498 & 0.691 & 0.512 & 0.701 & 0.506 & 0.707 & 0.511 \\
    & 336 & \textbf{0.704} & \textbf{0.497} & \underline{0.711} & 0.515 & \textbf{0.704} & 0.515 & 0.732 & \underline{0.503} & 0.713 & 0.510 & 0.715 & 0.510 & 0.716 & 0.529 & 0.722 & 0.519 & 0.785 & 0.537 \\
    & 720 & 0.764 & 0.522 & 0.770 & 0.538 & \textbf{0.747} & \underline{0.518} & 0.783 & \textbf{0.515} & 0.766 & 0.537 & 0.756 & 0.534 & 0.765 & 0.556 & 0.777 & 0.545 & \underline{0.755} & 0.527 \\
    \cmidrule{2-20}
    & Avg  & \underline{0.695} & \textbf{0.491} & 0.704 & 0.508 & \textbf{0.694} & 0.504 & 0.722 & \underline{0.495} & 0.705 & 0.506 & 0.706 & 0.506 & 0.705 & 0.522 & 0.713 & 0.514 & 0.726 & 0.515 \\
    \midrule
    \multirow{5}{*}{\rotatebox{90}{AQWan}}
    & 96  & \textbf{0.711} & \textbf{0.446} & 0.745 & 0.470 & 0.750 & 0.465 & 0.761 & \underline{0.458} & 0.747 & 0.470 & \underline{0.744} & 0.468 & 0.756 & 0.481 & 0.758 & 0.475 & 0.791 & 0.488 \\
    & 192 & \underline{0.773} & \textbf{0.473} & 0.792 & 0.491 & \textbf{0.762} & 0.479 & 0.801 & \underline{0.478} & 0.787 & 0.486 & 0.804 & 0.488 & 0.800 & 0.502 & 0.809 & 0.496 & 0.779 & 0.490 \\
    & 336 & \textbf{0.799} & \textbf{0.486} & 0.819 & 0.503 & \underline{0.802} & 0.504 & 0.821 & \underline{0.488} & 0.814 & 0.497 & 0.813 & 0.500 & 0.823 & 0.516 & 0.830 & 0.508 & 0.814 & 0.505 \\
    & 720 & \underline{0.869} & 0.514 & 0.890 & 0.533 & \textbf{0.830} & \underline{0.511} & 0.888 & \textbf{0.506} & 0.889 & 0.529 & 0.878 & 0.522 & 0.891 & 0.548 & 0.906 & 0.538 & \underline{0.869} & 0.519 \\
    \cmidrule{2-20}
    & Avg  & \underline{0.788} & \textbf{0.479} & 0.811 & 0.499 & \textbf{0.786} & 0.489 & 0.817 & \underline{0.482} & 0.809 & 0.495 & 0.809 & 0.494 & 0.817 & 0.511 & 0.825 & 0.504 & 0.813 & 0.500 \\
    \midrule
    \multirow{5}{*}{\rotatebox{90}{CzeLan}}
    & 96  & \textbf{0.170} & \textbf{0.208} & 0.183 & 0.251 & 0.581 & 0.443 & \underline{0.172} & \underline{0.213} & 0.177 & 0.239 & 0.175 & 0.230 & 0.211 & 0.289 & 0.178 & 0.229 & 0.176 & 0.237 \\
    & 192 & \textbf{0.200} & \textbf{0.231} & 0.208 & 0.271 & 0.705 & 0.503 & 0.207 & \underline{0.236} & \underline{0.201} & 0.257 & 0.206 & 0.254 & 0.252 & 0.323 & 0.210 & 0.252 & 0.215 & 0.279 \\
    & 336 & \underline{0.228} & \textbf{0.257} & 0.243 & 0.302 & 0.971 & 0.596 & 0.240 & \underline{0.262} & 0.232 & 0.282 & 0.230 & 0.277 & 0.317 & 0.366 & 0.243 & 0.280 & \textbf{0.224} & 0.288 \\
    & 720 & \underline{0.262} & \textbf{0.286} & 0.273 & 0.335 & 1.566 & 0.762 & 0.288 & \underline{0.298} & \textbf{0.261} & 0.311 & \underline{0.262} & 0.309 & 0.358 & 0.392 & 0.284 & 0.317 & 0.282 & 0.337 \\
    \cmidrule{2-20}
    & Avg  & \textbf{0.215} & \textbf{0.245} & 0.226 & 0.289 & 0.955 & 0.576 & 0.226 & \underline{0.252} & \underline{0.217} & 0.272 & 0.218 & 0.267 & 0.284 & 0.342 & 0.228 & 0.269 & 0.224 & 0.285 \\
    \midrule
    \multirow{5}{*}{\rotatebox{90}{ZafNoo}}
    & 96  & 0.437 & \textbf{0.389} & 0.444 & 0.426 & \textbf{0.432} & 0.419 & 0.435 & \underline{0.391} & 0.439 & 0.408 & 0.441 & 0.396 & \underline{0.434} & 0.411 & 0.446 & 0.410 & 0.479 & 0.424 \\
    & 192 & 0.498 & \underline{0.429} & 0.498 & 0.456 & \textbf{0.432} & \textbf{0.419} & 0.501 & 0.432 & 0.505 & 0.443 & 0.498 & 0.444 & \underline{0.484} & 0.444 & 0.503 & 0.447 & 0.491 & 0.446 \\
    & 336 & 0.539 & \textbf{0.453} & 0.530 & 0.480 & \underline{0.521} & 0.469 & 0.551 & \underline{0.461} & 0.555 & 0.473 & 0.543 & 0.466 & \textbf{0.518} & 0.464 & 0.544 & 0.470 & 0.551 & 0.479 \\
    & 720 & 0.588 & \textbf{0.483} & 0.574 & 0.499 & \textbf{0.543} & \textbf{0.483} & 0.596 & \textbf{0.483} & 0.591 & 0.501 & 0.588 & 0.498 & \underline{0.548} & \underline{0.486} & 0.595 & 0.504 & 0.627 & 0.511 \\
    \cmidrule{2-20}
    & Avg  & 0.515 & \textbf{0.438} & 0.511 & 0.465 & \textbf{0.482} & 0.447 & 0.520 & \underline{0.441} & 0.522 & 0.456 & 0.517 & 0.451 & \underline{0.496} & 0.451 & 0.522 & 0.457 & 0.537 & 0.465 \\
    \midrule
    \bottomrule
    \end{tabular}
    }
\end{table*}

\begin{table*}[ht]
    \centering
    \caption{\textbf{Full results of multivariate long-term forecasting with a fine-tuned set of hyperparameters.}Best results in \textbf{bold}, second-best \underline{underlined}. \acs{fnf} consistently outperforms baselines, both in \ac {mse} and \ac {mae}.}
    \label{tab:tuned_results}
    \scalebox{0.6}{
    \begin{tabular}{@{}lc cc cc cc cc cc cc cc cc cc@{}}
    \toprule
    \multicolumn{2}{c}{Models} & \multicolumn{2}{c}{\ac{fnf}} & \multicolumn{2}{c}{PatchTST} & \multicolumn{2}{c}{Crossformer} & \multicolumn{2}{c}{Pathformer} & \multicolumn{2}{c}{iTransformer} & \multicolumn{2}{c}{TimeMixer} & \multicolumn{2}{c}{DLinear} & \multicolumn{2}{c}{NLinear} & \multicolumn{2}{c}{TimesNet} \\
    \midrule
    \multicolumn{2}{c}{Metrics} & \ac{mse} & \ac{mae} & \ac{mse} & \ac{mae} & \ac{mse} & \ac{mae} & \ac{mse} & \ac{mae} & \ac{mse} & \ac{mae} & \ac{mse} & \ac{mae} & \ac{mse} & \ac{mae} & \ac{mse} & \ac{mae} & \ac{mse} & \ac{mae} \\
    \midrule
    \multirow{5}{*}{\rotatebox{90}{ETTh1}}
    & 96  & \textbf{0.355} & \textbf{0.388} & 0.376 & 0.396 & 0.405 & 0.426 & 0.372 & \underline{0.392} & 0.386 & 0.405 & 0.372 & 0.401 & \underline{0.371} & \underline{0.392} & 0.372 & 0.393 & 0.389 & 0.412 \\
    & 192 & \textbf{0.396} & \textbf{0.412} & \underline{0.399} & 0.416 & 0.413 & 0.442 & 0.408 & 0.415 & 0.424 & 0.440 & 0.413 & 0.430 & 0.404 & \underline{0.413} & 0.405 & \underline{0.413} & 0.440 & 0.443 \\
    & 336 & \underline{0.425} & \textbf{0.425} & \textbf{0.418} & 0.432 & 0.442 & 0.460 & 0.438 & 0.434 & 0.449 & 0.460 & 0.438 & 0.450 & 0.434 & 0.435 & 0.429 & \underline{0.427} & 0.482 & 0.465 \\
    & 720 & \textbf{0.436} & \textbf{0.445} & \underline{0.450} & 0.469 & 0.550 & 0.539 & \underline{0.450} & 0.463 & 0.495 & 0.487 & 0.486 & 0.484 & 0.469 & 0.489 & \textbf{0.436} & \underline{0.452} & 0.525 & 0.501 \\
    \cmidrule{2-20}
    & Avg  & \textbf{0.403} & \textbf{0.417} & \underline{0.410} & 0.428 & 0.452 & 0.466 & 0.417 & 0.426 & 0.438 & 0.448 & 0.427 & 0.441 & 0.419 & 0.432 & \underline{0.410} & \underline{0.421} & 0.459 & 0.455 \\
    \midrule
    \multirow{5}{*}{\rotatebox{90}{ETTh2}}
    & 96  & \textbf{0.266} & \textbf{0.326} & 0.277 & 0.339 & 0.611 & 0.557 & 0.279 & \underline{0.336} & 0.297 & 0.348 & 0.281 & 0.351 & 0.302 & 0.368 & \underline{0.275} & 0.338 & 0.319 & 0.363 \\
    & 192 & \textbf{0.331} & \textbf{0.371} & 0.345 & 0.381 & 0.810 & 0.651 & 0.345 & 0.380 & 0.372 & 0.403 & 0.349 & 0.387 & 0.405 & 0.433 & \underline{0.336} & \underline{0.379} & 0.411 & 0.416 \\
    & 336 & \textbf{0.336} & \textbf{0.388} & 0.368 & 0.404 & 0.928 & 0.698 & 0.378 & 0.408 & 0.388 & 0.417 & 0.366 & 0.413 & 0.496 & 0.490 & \underline{0.362} & \underline{0.403} & 0.415 & 0.443 \\
    & 720 & \textbf{0.378} & \textbf{0.418} & 0.397 & \underline{0.432} & 1.094 & 0.775 & 0.437 & 0.455 & 0.424 & 0.444 & 0.401 & 0.436 & 0.766 & 0.622 & \underline{0.396} & 0.437 & 0.429 & 0.445 \\
    \cmidrule{2-20}
    & Avg  & \textbf{0.327} & \textbf{0.375} & 0.346 & \underline{0.389} & 0.860 & 0.670 & 0.359 & 0.394 & 0.370 & 0.403 & 0.349 & 0.396 & 0.492 & 0.478 & \underline{0.342} & \underline{0.389} & 0.393 & 0.416 \\
    \midrule
    \multirow{5}{*}{\rotatebox{90}{ETTm1}}
    & 96  & \textbf{0.286} & \textbf{0.330} & \underline{0.290} & 0.343 & 0.310 & 0.361 & \underline{0.290} & \underline{0.335} & 0.300 & 0.353 & 0.293 & 0.345 & 0.299 & 0.343 & 0.301 & 0.343 & 0.377 & 0.398 \\
    & 192 & \textbf{0.324} & \textbf{0.356} & \underline{0.329} & 0.368 & 0.363 & 0.402 & 0.337 & \underline{0.363} & 0.341 & 0.380 & 0.335 & 0.372 & \underline{0.334} & 0.364 & 0.337 & 0.365 & 0.405 & 0.411 \\
    & 336 & \textbf{0.358} & \textbf{0.376} & \underline{0.360} & 0.390 & 0.408 & 0.430 & 0.374 & \underline{0.384} & 0.374 & 0.394 & 0.368 & 0.386 & \underline{0.365} & \underline{0.384} & 0.371 & \underline{0.384} & 0.443 & 0.437 \\
    & 720 & 0.423 & \textbf{0.410} & \textbf{0.416} & 0.422 & 0.777 & 0.637 & 0.428 & 0.416 & 0.429 & 0.430 & 0.426 & 0.417 & \underline{0.418} & \underline{0.415} & 0.426 & \underline{0.415} & 0.495 & 0.464 \\
    \cmidrule{2-20}
    & Avg  & \textbf{0.347} & \textbf{0.368} & \underline{0.348} & 0.380 & 0.464 & 0.457 & 0.357 & \underline{0.374} & 0.361 & 0.389 & 0.355 & 0.380 & 0.354 & 0.376 & 0.358 & 0.376 & 0.430 & 0.427 \\
    \midrule
    \multirow{5}{*}{\rotatebox{90}{ETTm2}}
    & 96  & \textbf{0.158} & \textbf{0.244} & 0.165 & 0.254 & 0.263 & 0.359 & 0.164 & \underline{0.250} & 0.175 & 0.266 & 0.165 & 0.256 & 0.164 & 0.255 & \underline{0.163} & 0.252 & 0.190 & 0.266 \\
    & 192 & \textbf{0.214} & \textbf{0.282} & 0.221 & 0.292 & 0.361 & 0.425 & 0.219 & \underline{0.288} & 0.242 & 0.312 & 0.225 & 0.298 & 0.224 & 0.304 & \underline{0.218} & 0.290 & 0.251 & 0.308 \\
    & 336 & \textbf{0.267} & \textbf{0.317} & 0.275 & 0.325 & 0.469 & 0.496 & \textbf{0.267} & \underline{0.319} & 0.282 & 0.337 & 0.277 & 0.332 & 0.277 & 0.337 & \underline{0.273} & 0.326 & 0.322 & 0.350 \\
    & 720 & \textbf{0.360} & \underline{0.379} & \textbf{0.360} & 0.380 & 1.263 & 0.857 & \underline{0.361} & \textbf{0.377} & 0.375 & 0.394 & \textbf{0.360} & 0.387 & 0.371 & 0.401 & \underline{0.361} & 0.382 & 0.414 & 0.403 \\
    \cmidrule{2-20}
    & Avg  & \textbf{0.249} & \textbf{0.305} & 0.255 & 0.312 & 0.589 & 0.534 & 0.252 & \underline{0.308} & 0.268 & 0.327 & 0.256 & 0.318 & 0.259 & 0.324 & 0.253 & 0.312 & 0.294 & 0.331 \\
    \midrule
    \multirow{5}{*}{\rotatebox{90}{Electricity}}
    & 96  & \textbf{0.127} & \textbf{0.219} & \underline{0.133} & 0.233 & 0.135 & 0.237 & 0.135 & \underline{0.222} & 0.134 & 0.230 & 0.153 & 0.256 & 0.140 & 0.237 & 0.141 & 0.236 & 0.164 & 0.267 \\
    & 192 & \textbf{0.144} & \textbf{0.235} & \underline{0.150} & \underline{0.248} & 0.160 & 0.262 & 0.157 & 0.253 & 0.154 & 0.250 & 0.168 & 0.269 & 0.154 & 0.250 & 0.155 & \underline{0.248} & 0.180 & 0.280 \\
    & 336 & \textbf{0.157} & \textbf{0.249} & \underline{0.168} & 0.267 & 0.182 & 0.282 & 0.170 & 0.267 & 0.169 & 0.265 & 0.189 & 0.291 & 0.169 & 0.268 & 0.171 & \underline{0.264} & 0.190 & 0.292 \\
    & 720 & \textbf{0.184} & \textbf{0.274} & 0.202 & 0.295 & 0.246 & 0.337 & 0.211 & 0.302 & \underline{0.194} & \underline{0.288} & 0.228 & 0.320 & 0.204 & 0.301 & 0.210 & 0.297 & 0.209 & 0.307 \\
    \cmidrule{2-20}
    & Avg  & \textbf{0.153} & \textbf{0.244} & 0.163 & 0.260 & 0.180 & 0.279 & 0.168 & 0.261 & \underline{0.162} & \underline{0.258} & 0.184 & 0.284 & 0.166 & 0.264 & 0.169 & 0.261 & 0.185 & 0.286 \\
    \midrule
    \multirow{5}{*}{\rotatebox{90}{Weather}}
    & 96  & 0.149 & \textbf{0.188} & 0.149 & 0.196 & \textbf{0.146} & 0.212 & \underline{0.148} & \underline{0.195} & 0.157 & 0.207 & 0.147 & 0.198 & 0.170 & 0.230 & 0.179 & 0.222 & 0.170 & 0.219 \\
    & 192 & 0.195 & \textbf{0.231} & 0.193 & 0.240 & 0.195 & 0.261 & \textbf{0.191} & \underline{0.235} & 0.200 & 0.248 & \underline{0.192} & 0.243 & 0.212 & 0.267 & 0.218 & 0.261 & 0.222 & 0.264 \\
    & 336 & 0.245 & \textbf{0.271} & \underline{0.244} & 0.281 & 0.268 & 0.325 & \textbf{0.243} & 0.274 & 0.252 & 0.287 & 0.247 & 0.284 & 0.257 & 0.305 & 0.266 & 0.296 & 0.293 & 0.310 \\
    & 720 & \textbf{0.314} & \textbf{0.324} & \textbf{0.314} & 0.332 & 0.330 & 0.380 & \underline{0.318} & \underline{0.326} & 0.320 & 0.336 & \underline{0.318} & 0.330 & \underline{0.318} & 0.356 & 0.334 & 0.344 & 0.360 & 0.355 \\
    \cmidrule{2-20}
    & Avg  & \textbf{0.225} & \textbf{0.253} & \textbf{0.225} & 0.262 & 0.234 & 0.294 & \textbf{0.225} & \underline{0.257} & 0.232 & 0.269 & \underline{0.226} & 0.263 & 0.239 & 0.289 & 0.249 & 0.280 & 0.261 & 0.287 \\
    \midrule
    \multirow{5}{*}{\rotatebox{90}{Traffic}}
    & 96  & \underline{0.365} & \textbf{0.231} & 0.379 & 0.271 & 0.514 & 0.282 & 0.384 & \underline{0.250} & \textbf{0.363} & 0.265 & 0.369 & 0.257 & 0.410 & 0.282 & 0.410 & 0.279 & 0.600 & 0.313 \\
    & 192 & \textbf{0.382} & \textbf{0.239} & 0.394 & 0.277 & 0.501 & 0.273 & 0.405 & \underline{0.257} & \underline{0.384} & 0.273 & 0.400 & 0.272 & 0.423 & 0.288 & 0.423 & 0.284 & 0.619 & 0.328 \\
    & 336 & \textbf{0.395} & \textbf{0.246} & 0.404 & 0.281 & 0.507 & 0.279 & 0.424 & \underline{0.265} & \underline{0.396} & 0.277 & 0.407 & 0.272 & 0.436 & 0.296 & 0.436 & 0.291 & 0.627 & 0.330 \\
    & 720 & \textbf{0.432} & \textbf{0.268} & \underline{0.442} & 0.302 & 0.571 & 0.301 & 0.452 & \underline{0.283} & 0.445 & 0.308 & 0.461 & 0.316 & 0.466 & 0.315 & 0.464 & 0.308 & 0.659 & 0.342 \\
    \cmidrule{2-20}
    & Avg  & \textbf{0.393} & \textbf{0.246} & 0.404 & 0.282 & 0.523 & 0.283 & 0.416 & \underline{0.263} & \underline{0.397} & 0.280 & 0.409 & 0.279 & 0.433 & 0.295 & 0.433 & 0.290 & 0.626 & 0.328 \\
    \midrule
    \multirow{5}{*}{\rotatebox{90}{AQShunyi}}
    & 96  & \textbf{0.632} & \textbf{0.463} & \underline{0.648} & 0.481 & 0.652 & 0.484 & 0.667 & \underline{0.472} & 0.650 & 0.479 & 0.654 & 0.483 & 0.651 & 0.492 & 0.653 & 0.486 & 0.658 & 0.488 \\
    & 192 & \underline{0.675} & \textbf{0.483} & 0.690 & 0.501 & \textbf{0.674} & 0.499 & 0.707 & \underline{0.491} & 0.693 & 0.498 & 0.700 & 0.498 & 0.691 & 0.512 & 0.701 & 0.506 & 0.707 & 0.511 \\
    & 336 & \textbf{0.699} & \textbf{0.497} & 0.711 & 0.515 & \underline{0.704} & 0.515 & 0.732 & \underline{0.503} & 0.713 & 0.510 & 0.715 & 0.510 & 0.716 & 0.529 & 0.722 & 0.519 & 0.785 & 0.537 \\
    & 720 & 0.758 & 0.525 & 0.770 & 0.538 & \textbf{0.747} & \underline{0.518} & 0.783 & \textbf{0.515} & 0.766 & 0.537 & 0.756 & 0.534 & 0.765 & 0.556 & 0.777 & 0.545 & \underline{0.755} & 0.527 \\
    \cmidrule{2-20}
    & Avg  & \textbf{0.691} & \textbf{0.492} & 0.704 & 0.508 & \underline{0.694} & 0.504 & 0.722 & \underline{0.495} & 0.705 & 0.506 & 0.706 & 0.506 & 0.705 & 0.522 & 0.713 & 0.514 & 0.726 & 0.515 \\
    \midrule
    \multirow{5}{*}{\rotatebox{90}{AQWan}}
    & 96  & \textbf{0.716} & \textbf{0.449} & 0.745 & 0.470 & 0.750 & 0.465 & 0.761 & \underline{0.458} & 0.747 & 0.470 & \underline{0.744} & 0.468 & 0.756 & 0.481 & 0.758 & 0.475 & 0.791 & 0.488 \\
    & 192 & \underline{0.769} & \textbf{0.473} & 0.792 & 0.491 & \textbf{0.762} & 0.479 & 0.801 & \underline{0.478} & 0.787 & 0.486 & 0.804 & 0.488 & 0.800 & 0.502 & 0.809 & 0.496 & 0.779 & 0.490 \\
    & 336 & \textbf{0.796} & \textbf{0.487} & 0.819 & 0.503 & \underline{0.802} & 0.504 & 0.821 & \underline{0.488} & 0.814 & 0.497 & 0.813 & 0.500 & 0.823 & 0.516 & 0.830 & 0.508 & 0.814 & 0.505 \\
    & 720 & \underline{0.867} & 0.516 & 0.890 & 0.533 & \textbf{0.830} & \underline{0.511} & 0.888 & \textbf{0.506} & 0.889 & 0.529 & 0.878 & 0.522 & 0.891 & 0.548 & 0.906 & 0.538 & 0.869 & 0.519 \\
    \cmidrule{2-20}
    & Avg  & \underline{0.787} & \textbf{0.481} & 0.811 & 0.499 & \textbf{0.786} & 0.489 & 0.817 & \underline{0.482} & 0.809 & 0.495 & 0.809 & 0.494 & 0.817 & 0.511 & 0.825 & 0.504 & 0.813 & 0.500 \\
    \midrule
    \multirow{5}{*}{\rotatebox{90}{CzeLan}}
    & 96  & \textbf{0.170} & \textbf{0.208} & 0.183 & 0.251 & 0.581 & 0.443 & \underline{0.172} & \underline{0.213} & 0.177 & 0.239 & 0.175 & 0.230 & 0.211 & 0.289 & 0.178 & 0.229 & 0.176 & 0.237 \\
    & 192 & \textbf{0.200} & \textbf{0.231} & 0.208 & 0.271 & 0.705 & 0.503 & 0.207 & \underline{0.236} & \underline{0.201} & 0.257 & 0.206 & 0.254 & 0.252 & 0.323 & 0.210 & 0.252 & 0.215 & 0.279 \\
    & 336 & \underline{0.228} & \textbf{0.257} & 0.243 & 0.302 & 0.971 & 0.596 & 0.240 & \underline{0.262} & 0.232 & 0.282 & 0.230 & 0.277 & 0.317 & 0.366 & 0.243 & 0.280 & \textbf{0.224} & 0.288 \\
    & 720 & \underline{0.262} & \textbf{0.286} & 0.273 & 0.335 & 1.566 & 0.762 & 0.288 & \underline{0.298} & \textbf{0.261} & 0.311 & \underline{0.262} & 0.309 & 0.358 & 0.392 & 0.284 & 0.317 & 0.282 & 0.337 \\
    \cmidrule{2-20}
    & Avg  & \textbf{0.215} & \textbf{0.245} & 0.226 & 0.289 & 0.955 & 0.576 & 0.226 & \underline{0.252} & \underline{0.217} & 0.272 & 0.218 & 0.267 & 0.284 & 0.342 & 0.228 & 0.269 & 0.224 & 0.285 \\
    \midrule
    \multirow{5}{*}{\rotatebox{90}{ZafNoo}}
    & 96  & 0.437 & \textbf{0.389} & 0.444 & 0.426 & \textbf{0.432} & 0.419 & 0.435 & \underline{0.391} & 0.439 & 0.408 & 0.441 & 0.396 & \underline{0.434} & 0.411 & 0.446 & 0.410 & 0.479 & 0.424 \\
    & 192 & 0.498 & \underline{0.429} & 0.498 & 0.456 & \textbf{0.432} & \textbf{0.419} & 0.501 & 0.432 & 0.505 & 0.443 & 0.498 & 0.444 & \underline{0.484} & 0.444 & 0.503 & 0.447 & 0.491 & 0.446 \\
    & 336 & 0.539 & \textbf{0.453} & 0.530 & 0.480 & \underline{0.521} & 0.469 & 0.551 & \underline{0.461} & 0.555 & 0.473 & 0.543 & 0.466 & \textbf{0.518} & 0.464 & 0.544 & 0.470 & 0.551 & 0.479 \\
    & 720 & 0.588 & \textbf{0.483} & 0.574 & 0.499 & \textbf{0.543} & \textbf{0.483} & 0.596 & \textbf{0.483} & 0.591 & 0.501 & 0.588 & 0.498 & \underline{0.548} & \underline{0.486} & 0.595 & 0.504 & 0.627 & 0.511 \\
    \cmidrule{2-20}
    & Avg  & 0.515 & \textbf{0.438} & 0.511 & 0.465 & \textbf{0.482} & 0.447 & 0.520 & \underline{0.441} & 0.522 & 0.456 & 0.517 & 0.451 & \underline{0.496} & 0.451 & 0.522 & 0.457 & 0.537 & 0.465 \\
    \midrule
    \bottomrule
    \end{tabular}
    }
\end{table*}

\end{document}